%% file: iclr2027_conference.tex
\documentclass{article} 
\usepackage{iclr2027_conference,times}

\input{math_commands.tex}

\usepackage{hyperref}
\usepackage{url}
\usepackage[pdftex]{graphicx}
\usepackage{pdfcomment}
\usepackage{booktabs}
\usepackage{subcaption}
\usepackage{pifont}
\usepackage{bbding}
\usepackage{amssymb}
\usepackage{multirow}
\usepackage{algorithm}
\usepackage{algorithmic}
\usepackage{colortbl}
\usepackage{multirow}
\usepackage{makecell}
\usepackage{array}
\usepackage[table]{xcolor}
\usepackage{tablefootnote}
\usepackage{tabularx}
\usepackage[raggedrightboxes]{ragged2e}
\definecolor{softblue}{rgb}{0.3, 0.5, 0.8} 
\definecolor{softgreen}{HTML}{4472C4}

\definecolor{lightblue}{HTML}{89CFF0}
\hypersetup{
  colorlinks=true,        
  linkcolor=softblue,    
  citecolor=green,    
  urlcolor=lightblue,     
  filecolor=lightblue,    
  pdfborder={0 0 0}       
}

\title{CineMem: Attention-Aligned Fine-grained Memory Retrieval for Multi-Shot Long Video Generation}

\author{
Xiaoxue Wu$^{1,2*}$, Xinyu Gao$^{3}$, Fang Yu$^{1,4}$, Yu Qiao$^{1\dag}$, Xinyuan Chen$^{1\dag}$ \\
$^{1}$Shanghai Artificial Intelligence Laboratory  \\
$^{2}$Fudan University\\
$^{3}$Shanghai Jiao Tong University \\
$^{4}$East China Normal University \\
}

\newcommand{\methodname}{CineMem}

\iclrfinalcopy 
\begin{document}

\maketitle

\begin{figure}[h]
\begin{center}
    \includegraphics[width=1.0\linewidth]{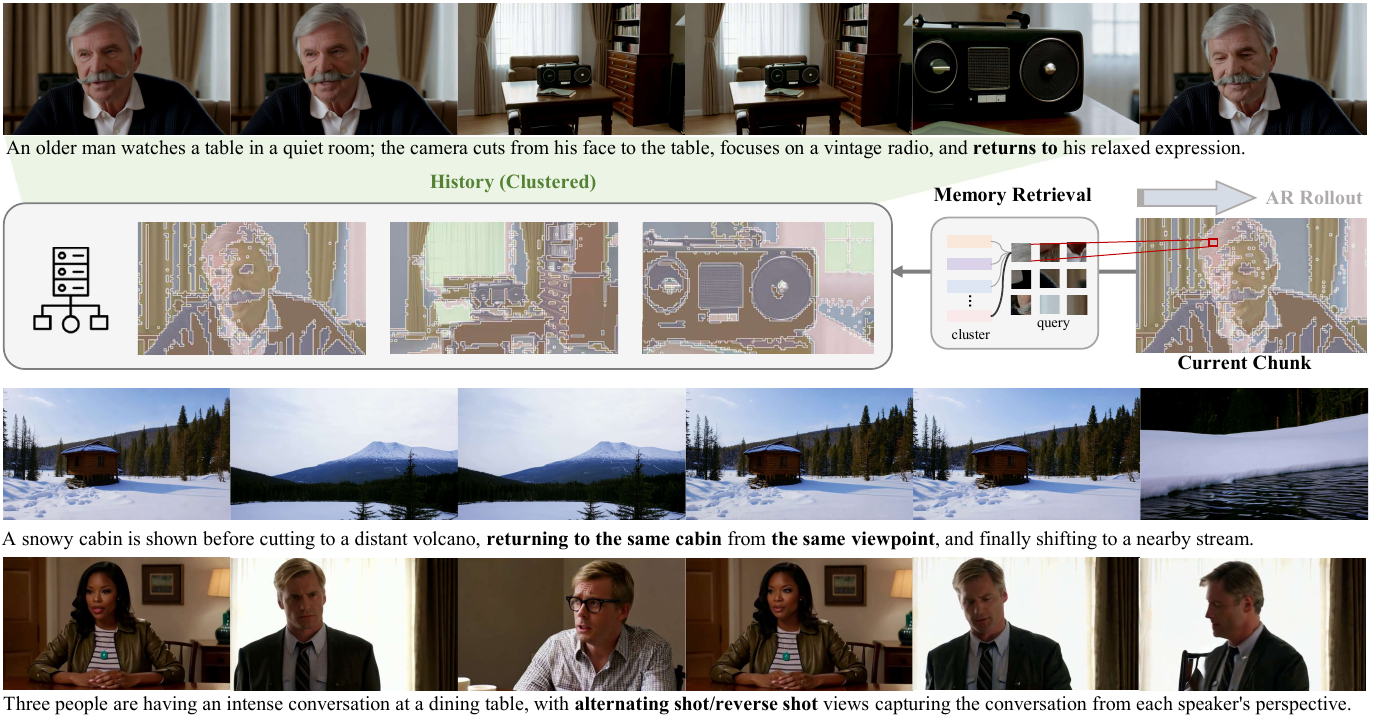}
\end{center}
\caption{15-second multi-shot videos autoregressively generated by \textbf{\methodname{}}. By combining clustered memory with fine-grained query-adaptive retrieval,  \methodname{} maintains coherent semantic flow across shot transitions, particularly in revisitation scenarios. Project page: 
\url{https://uknowsth.github.io/CineMem/}
}
\label{fig:teaser}
\end{figure}

\renewcommand{\thefootnote}{}
\footnotetext{\textsuperscript{*}Work done during an internship at Shanghai AI Laboratory.\\   
\quad\quad \textsuperscript{\dag}Corresponding authors.}
\renewcommand{\thefootnote}{\arabic{footnote}}

\input{sec/0_abstract}
\input{sec/1_intro}
\input{sec/2_related}
\input{sec/3_method}
\input{sec/4_experiment}
\input{sec/5_conclusion}

\bibliography{iclr2027_conference}
\bibliographystyle{iclr2027_conference}

\clearpage
\appendix


\begingroup
\hypersetup{linkcolor=black}
\setlength{\baselineskip}{0.9\baselineskip}
\tableofcontents
\endgroup





\input{sec/A_memory_access}
\input{sec/B_Methodology}

\input{sec/C_efficiency}
\input{sec/D_evaluation}
\input{sec/E_Results}

\input{sec/F_limit}
\input{sec/G_future}

\end{document}

%% file: math_commands.tex
\usepackage{amsmath,amsfonts,bm}

\def\eqref#1{equation~\ref{#1}}

\def\1{\bm{1}}

\DeclareMathAlphabet{\mathsfit}{\encodingdefault}{\sfdefault}{m}{sl}
\SetMathAlphabet{\mathsfit}{bold}{\encodingdefault}{\sfdefault}{bx}{n}



%% file: sec/0_abstract.tex
\begin{abstract}
Autoregressive video generation relies on memory retrieval to model long-range dependencies. This capability becomes particularly critical for multi-shot generation, where models must reconcile substantial visual changes across shot transitions with persistent semantic continuity.
Existing methods tend to perform memory retrieval at the frame level and share the same historical context across all query tokens, overlooking the heterogeneous information needs and leading to redundant retrieval and inefficient use of historical memory.
To bridge this gap, we introduce \methodname{}, an attention-aligned fine-grained memory retrieval framework that combines online memory organization with query-adaptive retrieval for autoregressive multi-shot video generation.
Specifically, guided by intrinsic memory-access patterns revealed through attention analysis, we propose an online-clustered KV cache as a memory organization strategy over all historical tokens, enabling semantic-aware memory grouping and more precise retrieval. 
In addition, we introduce a query-level retrieval mechanism that provides broader access to historical information under a fixed retrieval budget, thereby improving long-term memory utilization. Together, these designs enable the multi-shot video generator to preserve long-range consistency across shot transitions, particularly in revisitation scenarios, while approximating full-memory access under a constrained memory budget.
To facilitate evaluation, we further introduce a suite of task-specific, VLM-assisted metrics for faithful inter-shot consistency assessment. Extensive experiments demonstrate the effectiveness of our framework.
\end{abstract}

%% file: sec/1_intro.tex
\section{Introduction}
\label{sec:intro}

Building on bidirectional architectures and extensive pre-training, recent diffusion models \citep{stablevideodiffusion,ho2020denoising,rombach2022high,song2020score} deliver impressive visual quality, but at substantial computational cost for long-duration generation. 
Autoregressive generation offers a natural alternative for streaming video synthesis, typically combining a sliding-window mechanism with memory routing to maintain long-range consistency and stable generation \citep{huang2026self,yin2025slow}. This capability is particularly important for multi-shot video generation \citep{guo2025longcontexttuning,luo2026shotstream}, which aims to synthesize coherent, film-like narratives through frequent shot transitions accompanied by substantial viewpoint changes. Maintaining a continuous and coherent narrative flow across discrete shot boundaries naturally calls for a well-designed memory retrieval strategy, especially 
when previously observed subjects, scenes, or viewpoints reappear after intervening shots.

Existing memory mechanisms for causal video generation remain generally coarse-grained. Basic sliding-window strategies are widely adopted in single-shot causal models \citep{huang2026self,yang2025longlive}, inevitably discarding distant information as autoregressive rollout proceeds. In multi-shot scenarios, some methods preserve temporally predefined anchors as persistent memory \citep{chen2026longlive2}, such as the first frames of individual shots. 
These designs follow fixed routing patterns determined by temporal structures rather than the model's intrinsic memory-access patterns, limiting their ability to fully exploit historical information.
More recent studies introduce retrieval-based memory strategies \citep{meng2026causalcine,luo2026shotstream}, which typically augment fixed sinks and adjacent context with frame-level routing.
The resulting routing decision is shared across all query tokens.
Although such designs allow content-aware access to historical cues, they still rely on coarse temporal grouping and overlook heterogeneous memory requirements across queries.
Consequently, long-horizon autoregressive generation remains constrained by frame-level historical conditioning, leaving much of the available memory underutilized.

In this work, we explore how an autoregressive multi-shot video model can retrieve long-term historical information under a constrained memory budget by leveraging its intrinsic memory-access patterns.
Rather than merely preserving adjacent-frame information for pixel-level continuity, memory in multi-shot scenarios should support the recovery of previously observed visual content after multiple shot transitions, a setting we refer to as revisitation. 
To this end, we propose \methodname{}, an attention-aligned fine-grained memory retrieval framework that combines online memory organization with query-adaptive retrieval for autoregressive multi-shot video generation. 
By adapting memory organization and retrieval granularity to attention-based selection patterns, \methodname{} enables more effective access to long-term history and improves long-range consistency across shot transitions, particularly in revisitation scenarios, as illustrated in Figure \ref{fig:teaser}.

To investigate the intrinsic memory-access patterns of autoregressive diffusion models, 
we first develop a training framework that converts a bidirectional video model into a long-duration autoregressive multi-shot generator operating in a chunk-by-chunk manner, while allowing each current chunk to attend to the full generation history. 
To stabilize long-horizon rollout, we further incorporate an error accumulation mitigation strategy during adaptation.
We then analyze the attention distributions between query tokens in the current chunk and all historical memory tokens. We observe a clear clustering structure among historical memory tokens and substantial variation in retrieval preferences across query tokens and attention heads. 
These observations motivate our designs for online memory organization and query-adaptive retrieval.

\begin{figure}[!t]
    \begin{center}
        \includegraphics[width=1.0\linewidth]{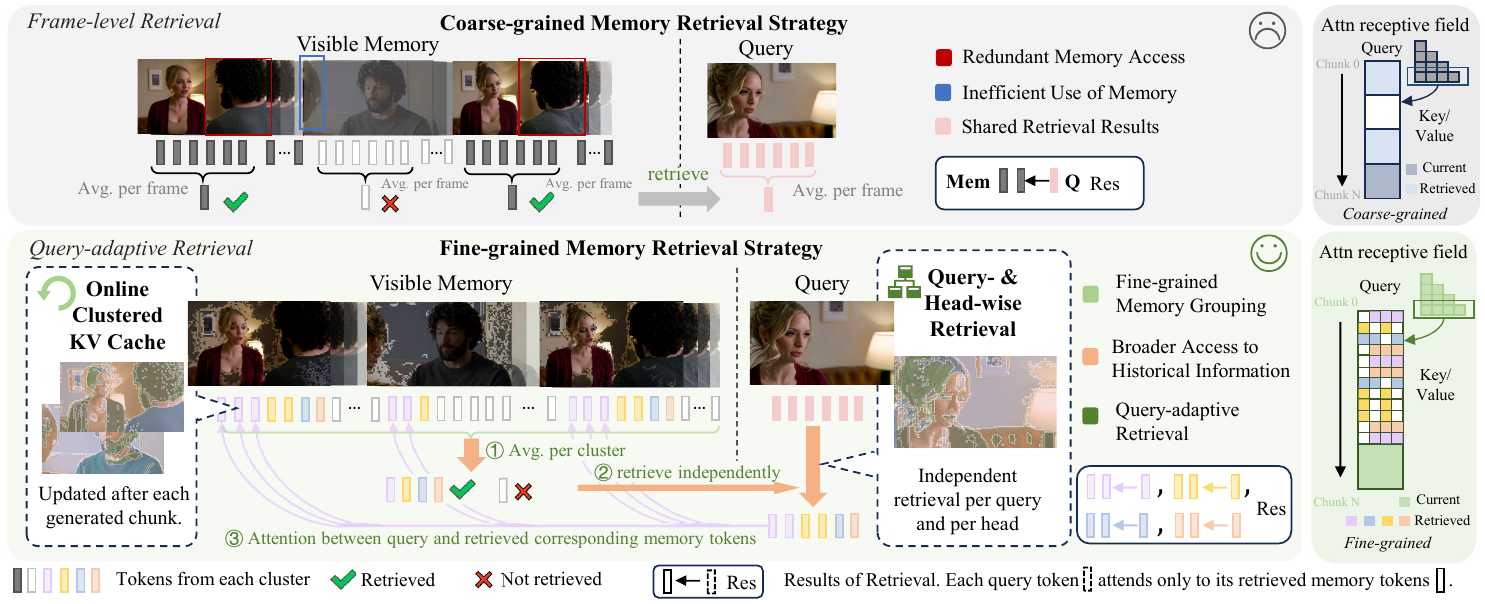}
    \end{center}
    \caption{\textbf{Overview of \methodname{}.} Existing autoregressive video generation methods tend to perform memory retrieval at the frame level, which may result in redundant retrieval, inefficient memory utilization, and overlooked heterogeneous information needs across query tokens. \methodname{} combines an online-clustered KV cache with query-adaptive retrieval to enable a finer-grained memory retrieval strategy. As illustrated by the attention receptive fields, \methodname{} provides broader (2/3 vs. 12/12) access to historical information under the same retrieval budget.} 
   \label{fig:overview}  
\end{figure}

Specifically, our framework comprises two complementary designs, as illustrated in Figure \ref{fig:overview}. First, we apply online clustering to the KV cache to organize historical memory throughout autoregressive rollout. As generation proceeds, historical tokens are grouped according to their semantic similarity, with each cluster serving as a retrieval unit. 
In contrast to conventional frame-level schemes that partition tokens by fixed frame membership regardless of content, our design forms content-aware groups, which alleviates the redundancy introduced by retrieving entire frames, while avoiding the high computational cost of treating every historical token as an individual memory unit.
Second, we introduce query-adaptive memory access in a query-wise and head-wise manner. Given their heterogeneous retrieval preferences, each query-head pair selects its most relevant clusters from memory. 
Under a fixed per-query memory budget, this individualized retrieval provides broader access to the generation history, yielding a larger effective attention receptive field without increasing the retrieval cost.
Together, these designs better match intrinsic memory-access patterns while balancing
retrieval granularity, historical coverage, and computational efficiency, enabling \methodname{} to maintain long-range consistency across frequent shot transitions.

For evaluation, we adopt a comprehensive protocol covering video-level quality, intra-shot consistency, and inter-shot consistency.  Since conventional similarity-based metrics can assign deceptively high inter-shot consistency scores to videos that fail to perform the intended shot transition, we further introduce a VLM-assisted evaluation protocol that jointly considers transition execution and semantic preservation. This provides a more faithful assessment of inter-shot consistency in multi-shot generation. Experimental results show that \methodname{} achieves superior consistency compared with existing autoregressive baselines while remaining competitive with large-scale bidirectional models.
Moreover, our attention-aligned retrieval designs are effective in a plug-in setting without additional fine-tuning, demonstrating their ability to better exploit historical information during autoregressive rollout. 
Overall, our work highlights the importance of fine-grained memory organization and retrieval for coherent long-horizon autoregressive multi-shot video generation.

%% file: sec/2_related.tex
\section{Related Work}
\label{sec:related}

\textbf{Autoregressive Video Generation and Memory.}
Diffusion-based autoregressive video generation enables streaming synthesis by progressively generating frames or chunks through autoregressive rollout \citep{gao2024vid,chen2024diffusion,liu2026realwonder,bruce2024genie,liu2026rolling,guo2025end}. 
Recent methods adapt pretrained bidirectional video models through post-training or distillation \citep{zhu2026causal,huang2026self,yin2025slow,cui2026self}, relying on fixed anchors, sliding windows, or coarse-grained memory retrieval to maintain long-range consistency \citep{ji2025memflow,hu2026longlive}. OminiMem \citep{zhao2026omnimem} further introduces a fine-grained head-adaptive retrieval framework, but primarily focuses on single-shot video generation. 
Meanwhile, fine-grained KV-cache organization has been explored in LLMs \citep{liu2025clusterkv,liu2026chunkkv,zhu2025sentencekv} mainly for inference acceleration in the text domain. 
In contrast, our work investigates fine-grained retrieval over visual memory with the primary goal of preserving long-range consistency, which we study in the more demanding setting of autoregressive multi-shot video generation.

\textbf{Multi-Shot Video Generation.}
Multi-shot video generation aims to synthesize sequences with discrete shot transitions while preserving narrative, subject, and scene consistency. Most existing approaches are built upon bidirectional video models and can be broadly categorized into two paradigms: generating individual shots separately and subsequently composing them \citep{he2023animateastory,xie2024dreamfactory,videogenofthought,long2024videostudio}, or directly modeling multi-shot sequences in an end-to-end manner \citep{guo2025longcontexttuning,qi2025mask2dit,wu2026cinetrans,wu2026shotdirector,meng2026holocine,wang2026multishotmaster,jia2026moga}. In autoregressive video generation, several methods enable dynamic content control by updating prompts during streaming synthesis, but primarily focus on single-shot generation \citep{yang2025longlive,ji2025memflow}. More recently, autoregressive approaches have begun to explore multi-shot generation. ShotStream \citep{luo2026shotstream} reformulates the problem as next-shot generation, while CausalCine introduces content-aware memory routing for accessing historical information \citep{meng2026causalcine}. Nevertheless, their memory retrieval remains relatively coarse-grained, limiting the effective use of long-term historical information.

%% file: sec/3_method.tex
\section{Method}
\label{sec:method}

In this section, we present the methodology and insights of \methodname{}. After introducing the preliminaries in Section \ref{sec:preliminary}, we describe our autoregressive adaptation framework in Section \ref{sec:training}, which enables the analysis of intrinsic memory-access patterns in Section \ref{sec:insight}. Guided by these insights, Sections \ref{sec:clusterKV} and \ref{sec:query} present our online-clustered KV cache and query-adaptive memory retrieval mechanism. Figure \ref{fig:method} provides an overview of \methodname{}.

\subsection{Preliminary}
\label{sec:preliminary}

\textbf{Flow-Matching.}
Flow matching learns a continuous velocity field along a probability path connecting noise and data samples \citep{lipman2022flow}. Given a clean latent $x_0$ and Gaussian noise $\epsilon$, the intermediate state is defined as $x_t=(1-t) x_0+t\epsilon$, and the velocity field is optimized by:
\begin{equation}
    \mathcal{L}_{\text{FM}} = \mathbb{E}_{t,x_0,\epsilon}[\Vert v_{\theta}(x_t,t) - (\epsilon - x_0) \Vert^2_2].
    \label{eq:fm}
\end{equation}
Modern video diffusion models typically parameterize such velocity fields with transformer-based architectures \citep{peebles2023scalable,vaswani2017attention}, where $v_{\theta}$ predicts the transport direction of video latents conditioned on text prompts or other conditions.

\textbf{Autoregressive Diffusion with Teacher Forcing.}
Built upon flow-matching video diffusion, autoregressive generation introduces causal dependency by generating each frame or chunk conditioned on previous history, formulated as $v_\theta(x_i^t,t,x_{<i},c)$. During training, teacher forcing replaces the generated history with ground-truth history:
\begin{equation}
    \mathcal{L}_{\text{AR}} = \mathbb{E}_{t,x,\epsilon}[\Vert v_{\theta}(x_i^t, t,x_{<i}^{gt},c) - (\epsilon_i - x_i) \Vert^2_2].
    \label{eq:tf}
\end{equation}
In multi-shot generation, the condition $c$ can be updated during autoregressive rollout to control shot transitions, resulting in substantial visual changes across shots while requiring temporal consistency.

\subsection{Long-Horizon Autoregressive Adaptation}
\label{sec:training}

\begin{figure}[!t]
    \begin{center}
        \includegraphics[width=1.0\linewidth]{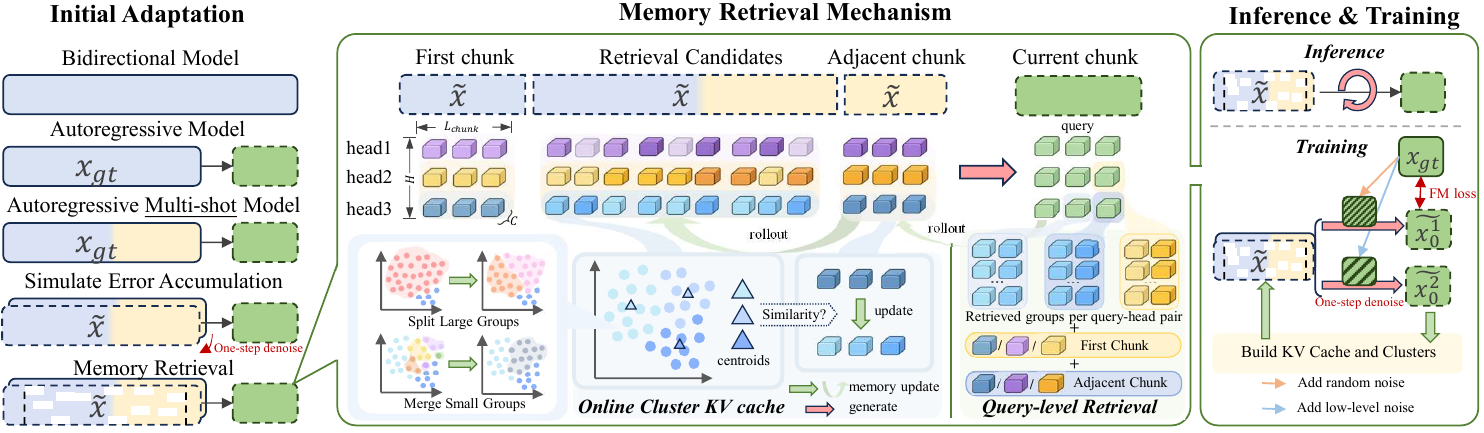}
    \end{center}
    \caption{\textbf{Our memory retrieval framework.} Through multi-stage initial adaptation, we convert a bidirectional model into an autoregressive multi-shot model with memory retrieval. During inference, queries access memory from the first chunk, adjacent chunk, and retrieved candidates via token-wise and head-wise retrieval. Previous adjacent chunks are progressively integrated into retrieval candidates through online KV-cache clustering with cluster balancing. During training, resampling ground truth frames provides memory inputs to alleviate error accumulation.} 
   \label{fig:method}  
\end{figure}

To analyze intrinsic memory-access patterns without introducing any explicit memory retrieval, we first adapt a bidirectional video model into a long-horizon autoregressive multi-shot generator with chunk rollout and prompt-controlled shot transitions, as illustrated in the left part of Figure \ref{fig:method}.

During adaptation, teacher forcing is first applied for autoregressive initialization. However, the resulting train-inference gap may cause error accumulation during long-horizon rollout and affect reliable memory analysis. Following \citet{guo2025end}, we mitigate this issue with resampled context training, where the ground-truth history is perturbed with low-level noise and denoised through a one-step rollout without gradient computation to obtain simulated inference history $\tilde{x}_{<i}$ that better approximates model-generated history.  Accordingly, Equation \ref{eq:tf} is reformulated as:
\begin{equation}
    \mathcal{L}_{\text{AR-resample}} = \mathbb{E}_{t,x,\epsilon}[\Vert v_{\theta}(x_i^t, t,\tilde{x}_{<i},c) - (\epsilon_i - x_i) \Vert^2_2].
    \label{eq:rf}
\end{equation}
This training strategy reduces train-inference discrepancy and improves long-horizon rollout stability, enabling the model to generate longer multi-shot videos with reduced error accumulation.

Meanwhile, to equip the model with multi-shot generation capability, we fine-tune the model on 15-second multi-shot videos. 
During training, a shot-level mask restricts each shot to attend only to the global description and its corresponding caption, enabling controllable shot transitions through prompt updates during autoregressive rollout.

\subsection{Memory Selection in Attention Module}
\label{sec:insight}

Based on the full-memory model, we analyze memory-access patterns during autoregressive rollout to guide the design of a fine-grained retrieval mechanism that best matches the model's intrinsic selection behavior under full-memory attention. 
While conventional retrieval schemes typically define the retrieval unit at the chunk-to-frame level, we revisit the fundamental token-to-token memory selection process inherent in attention, which motivates our fine-grained retrieval design. For each generated chunk, memory selection is essentially the process by which current tokens, serving as queries, select relevant historical memory tokens through their keys and values.
For a query, highly attended keys receive higher attention probabilities, which mathematically correspond to larger normalized dot-product similarities. We therefore use this signal to study token-wise selection relationships and reveal the underlying memory-selection patterns.

\begin{figure}[t]
  \centering
  \begin{subfigure}{0.64\linewidth}
   \includegraphics[width=1.0\linewidth]{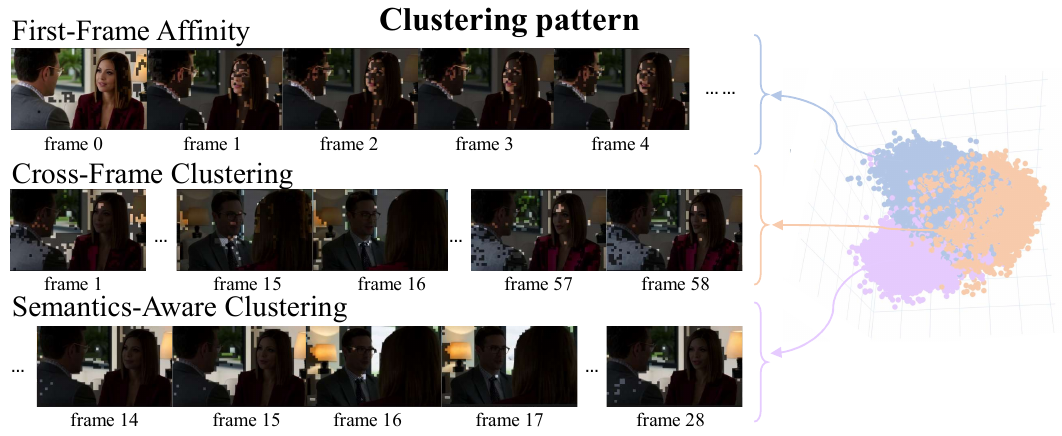}
   \caption{Clustering of full-memory tokens and representative patterns of selected clusters.}
   \label{fig:memory}      
  \end{subfigure}
  \hfill
  \begin{subfigure}{0.34\linewidth}
   \includegraphics[width=1.0\linewidth]{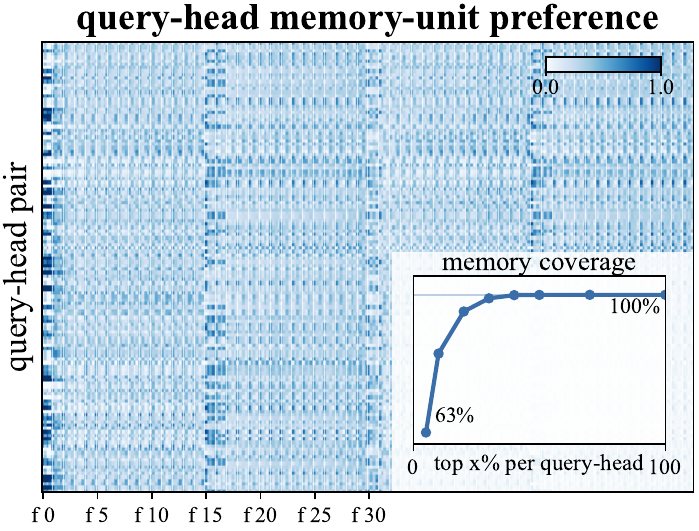}
   \caption{Heatmap of query-head memory preferences and memory coverage.} 
   \label{fig:query}
  \end{subfigure}
  \caption{\textbf{Intrinsic memory-selection patterns under the full-memory setting.} Memory tokens exhibit semantically meaningful clustering structures, while different query-head pairs show distinct preferences over memory tokens. Although each query-head pair selects a subset of the most relevant tokens, their union covers nearly the entire memory. See Appendix \ref{sec:sup_ana} for more details.} 
  \label{fig:visual}
\end{figure}

\textbf{Memory structure.} Efficient retrieval requires grouping, since treating each token as a separate group reduces retrieval to full-memory attention. Frame-level retrieval instead treats each frame as a group, implicitly assuming stronger correlations among tokens within the same frame. However, this assumption may not always hold, especially when a frame contains diverse content. To investigate this, we cluster the full-memory tokens and visualize the results in Figure \ref{fig:memory}. Clear clustering patterns emerge: highly similar tokens tend to form cross-frame clusters, while tokens within each cluster exhibit semantic consistency to some extent. We also observe stronger correlations among tokens from the first frame than among those from other frames, consistent with prior findings on the importance of the first latent sink \citep{chen2026first}. These observations motivate grouping memory by token similarity rather than frame membership. 
This is because frame-based grouping may introduce redundant information, as retrieving a group due to a few highly relevant tokens also includes many weakly related ones. Moreover, using a single centroid to represent semantically diverse tokens from the same frames can weaken their semantics and reduce retrieval precision.

\textbf{Heterogeneous memory selection.} 
Turning to the query side, each query in attention independently computes its similarity to all keys, naturally motivating query-level retrieval.
In practice, different queries and attention heads attend to distinct memory tokens.
We visualize these selection patterns in Figure \ref{fig:query} and observe clear heterogeneity across queries and heads. Even when each query-head pair selects only the top 30\% most relevant memory tokens, their overall memory receptive field covers nearly the full memory, which indicates the necessity of query-wise and head-wise retrieval.

\subsection{Online Clustered KV Cache}
\label{sec:clusterKV}

To enable fine-grained memory grouping, we introduce an online cluster KV cache. During autoregressive rollout, memory tokens are dynamically grouped according to their cosine similarity, such that semantically related tokens are assigned to the same cluster. 
This design leverages the semantic coherence within each cluster: tokens relevant to the same query are likely to share similar semantic information, reducing redundant retrieval while improving retrieval effectiveness.
To support fine-grained query-level retrieval, clustering is performed independently for each attention head.
We retain the first and latest chunks as memory anchors.
As rollout proceeds, expired adjacent chunks are progressively incorporated into retrieval candidates through online clustering. Specifically, newly introduced tokens are assigned to existing clusters based on centroid similarity, followed by cluster refinement.
To maintain balanced and stable clusters during online updates, we introduce several balancing strategies. In particular, over-sized and under-sized clusters are handled through explicit split and merge operations, as illustrated in Figure \ref{fig:method}, while cluster assignment is further stabilized by incorporating load penalties and assignment consistency terms into the assignment score, which is formulated as:
\begin{equation}
    z_i = \arg \max_c(\hat{k}_i^\top \mu_c - \lambda\phi(n_c)+\eta \mathbb{I}[c=z_i^{old}]),
    \label{eq:cluster}
\end{equation}
where $\hat{k}_i$ and $\mu_c$ denote the normalized key and centroid of cluster $c$, respectively; $\phi(n_c)$ is the cluster-size penalty, and the last term preserves previous assignments.

Through the online clustered KV cache, \methodname{} enables semantic-based grouping and establishes a context-aware memory organization strategy. By allowing cross-frame token associations, the proposed memory structure better exploits historical information and aligns with the model's intrinsic understanding of autoregressive generation, rather than reducing it to a frame-conditioned video extension process.
More details are provided in Appendix \ref{sec:sup_online_KV}.

\subsection{Query-level Retrieval}
\label{sec:query}

Considering the heterogeneous memory selection patterns discussed in Section \ref{sec:insight}, we design a query-level retrieval mechanism, specifically query-wise and head-wise retrieval, for fine-grained memory access. 
Conventional retrieval shares a single historical memory subset across all queries, which cannot capture diverse query and head preferences, causing redundant retrieval of irrelevant tokens and missing relevant historical information.
This motivates a more fine-grained retrieval unit. Instead of sharing, we allow each query-head pair to independently select relevant memory from the full history. Combined with the clustered KV cache in Section \ref{sec:clusterKV}, this design enables query-specific access to relevant memory clusters while maintaining the same retrieval budget, substantially expanding effective memory access and approximating full-memory attention under a constrained memory budget.
Based on query-centroid similarity, the retrieved memory is formulated as:
\begin{equation}
    \mathcal{M}_{q,h} = \mathcal{M}_{\text{first}} \cup 
    \{ (k_i,v_i) | z_i \in \mathcal{S}_{q,h} \}
    \cup \mathcal{M}_{\text{latest}}, \quad \mathcal{S}_{q,h}=\text{Retrieve}(q_h,\mathcal{C}_h),
    \label{eq:query}
\end{equation}
where $q_h$ denotes the query token of head $h$, $\mathcal{C}$ the clustered KV cache, and $\mathcal{S}_{q,h}$ the selected cluster set.
During training, we follow the error simulation mechanism introduced in Section \ref{sec:training} and construct KV caches from resampled ground-truth frames under low-level noise, which maintains consistency with the autoregressive adaptation setting, as illustrated in the right part of Figure \ref{fig:method}.
Further details are provided in Appendix \ref{sec:sup_query} and Appendix \ref{sec:sup_train}.

Together with the clustered KV cache, query-level retrieval better adapts memory access to the diverse information requirements of individual query tokens, leveraging query-specific preferences and head-specific representations to achieve effective memory retrieval.

%% file: sec/4_experiment.tex
\section{Experiment}
\label{sec:experiment}

In this section, we present the implementation details, evaluation settings, and experimental results of \methodname{}. Section \ref{sec:implementation} describes the implementation details and the dataset. Section \ref{sec:evaluation} introduces the evaluation protocol and results, and Section \ref{sec:ablation} presents ablation studies, which further validate the effectiveness of our design choices.
Additional results are provided in Appendix \ref{sec:sup_results}.

\subsection{Implementation details}
\label{sec:implementation}

\begin{figure}[!t]
  \centering
  \begin{subfigure}{0.95\linewidth}
   \includegraphics[width=1.0\linewidth]{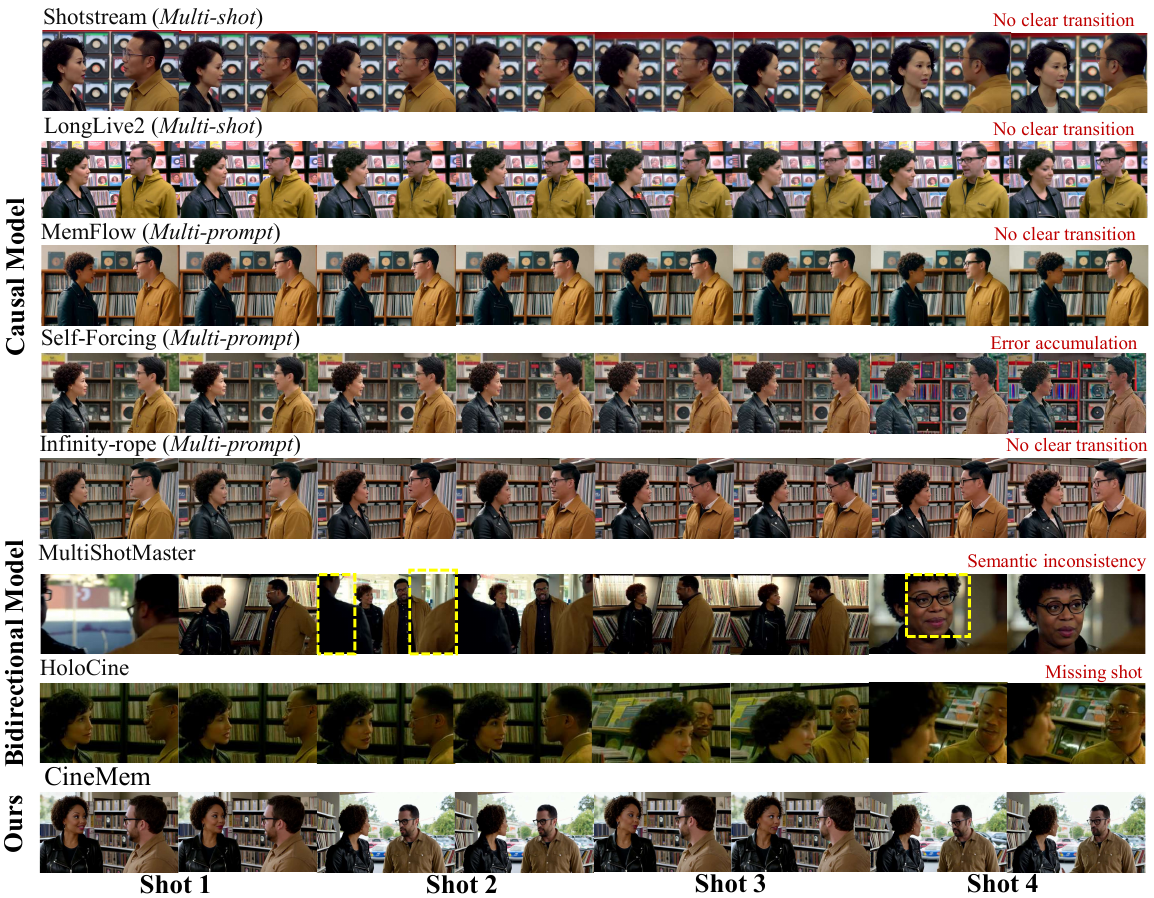}
  \end{subfigure}
  \hfill
  \begin{subfigure}{0.95\linewidth}
   \includegraphics[width=1.0\linewidth]{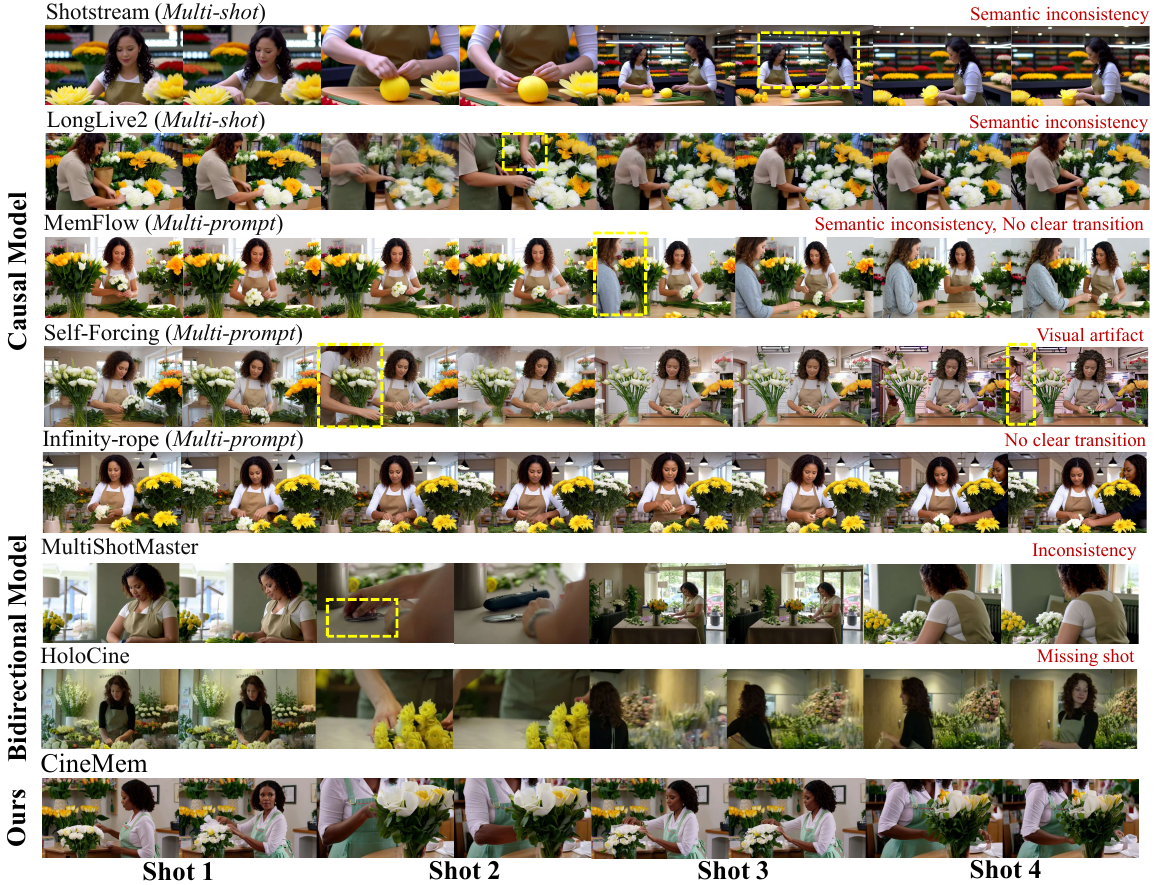}
  \end{subfigure}
  \caption{\textbf{Qualitative Results.} We compare \methodname{} with representative causal and bidirectional baselines on 15s multi-shot video generation. Representative examples illustrate the qualitative improvements of our method.} 
  \label{fig:results}
\end{figure}

We implement \methodname{} based on Wan2.1-T2V-1.3B \citep{wan2025wan} and train it on 16 NVIDIA H200 GPUs. For the autoregressive adaptation in Section \ref{sec:training}, we train for 20K steps after teacher forcing warmup, with the timestep shifting factor set to $s=0.6$ for low-level resampling following \citet{guo2025end}. We then train the memory retrieval model for 3K steps with a learning rate of $1\times10^{-5}$, using \texttt{flash\_attn\_varlen\_func} for efficient variable-length attention. The online clustered KV cache is implemented with K-means-style clustering, with target and maximum cluster sizes of 1,024 and 1,200 tokens, respectively. We retrieve 28 groups per query, approximately matching the attention budget of the original bidirectional model.

We compare the theoretical efficiency with full-history attention and frame-level retrieval in Table~\ref{tab:efficiency-theory}. The results show that \methodname{} reduces the overall computation relative to full-history attention while achieving a lower self-attention cost than frame-level retrieval, despite the additional overhead of online clustering. This overhead remains moderate relative to the dominant attention computation, enabling finer-grained and content-adaptive memory access at a competitive computational cost. See Appendix~\ref{sec:efficiency} for detailed analysis.

\textbf{Dataset.} We train on a 45K video dataset constructed from raw movie footage through shot-transition detection \citep{soucek2024transnetv2}, stitching \citep{girdhar2023imagebind}, aesthetic filtering \citep{LAIONAI}, OCR filtering, and detailed captioning \citep{achiam2023gpt}, resulting in a hierarchically annotated multi-shot video dataset.

\subsection{Evaluation and Results}
\label{sec:evaluation}


\begin{figure}[!t]
    \begin{center}
        \includegraphics[width=1.0\linewidth]{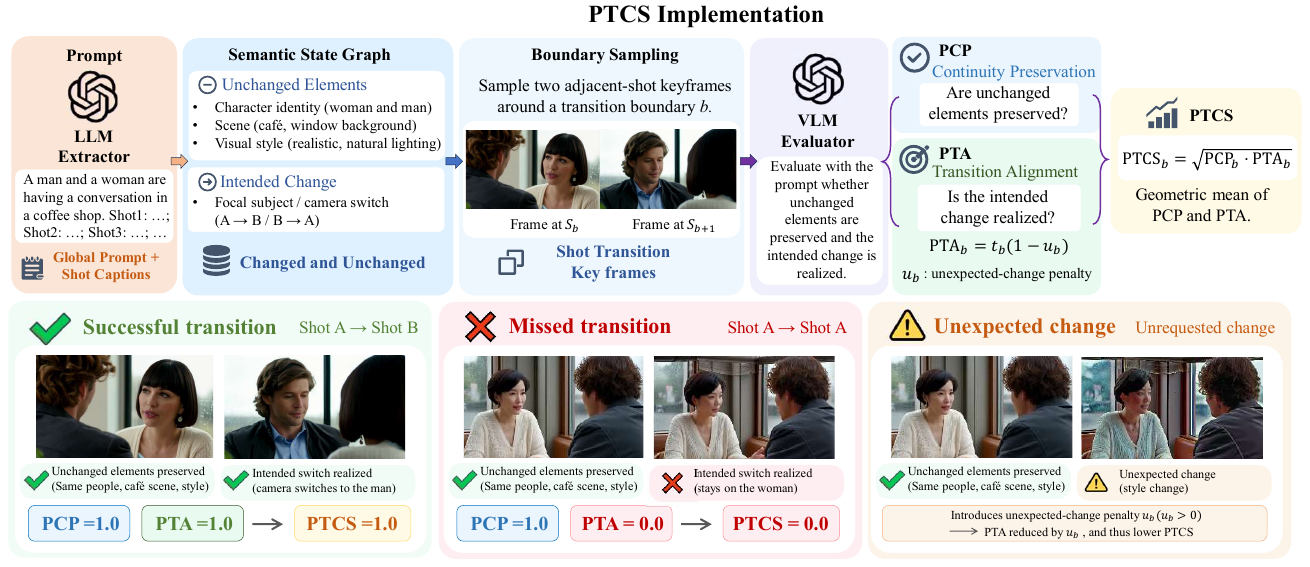  }
    \end{center}
    \caption{\textbf{Implementation of PTCS.} PTCS assigns low scores when either unchanged semantic elements are not preserved or intended changes are not correctly realized.} 
   \label{fig:PTCS}  
\end{figure}

\textbf{Metrics.} To comprehensively evaluate streaming multi-shot video generation, we construct an evaluation set of 100 prompts and establish an evaluation protocol covering video-level, intra-shot-level, and inter-shot-level metrics. Video-level metrics assess fundamental properties of generated videos, including visual aesthetics \citep{LAIONAI}, transition control \citep{soucek2024transnetv2}, and text alignment \citep{internvid}, while intra-shot metrics \citep{oquab2023dinov2, fu2023dreamsim} evaluate consistency within individual shots. More importantly, inter-shot evaluation measures semantic consistency across shot transitions.
However, existing similarity-based inter-shot metrics suffer from an inherent ambiguity: a high similarity score may indicate either strong transition consistency or simply the absence of a meaningful transition. Consequently, models with weak transition control may receive deceptively high consistency scores. 
To address this issue, we introduce a VLM-based metric \citep{achiam2023gpt}, the Prompt-conditioned Transition-Consistency Score (PTCS), as illustrated in Figure~\ref{fig:PTCS}. PTCS rewards transitions only when they are successfully executed while preserving the required semantic consistency.
PTCS consists of two complementary components: Prompt-conditioned Continuity Preservation (PCP) and Prompt-conditioned Transition Alignment (PTA). Given a hierarchical prompt, we first construct a semantic state graph to identify the semantic elements that should remain unchanged and those expected to change across the transition. PCP measures whether the required elements are consistently preserved before and after the transition, whereas PTA evaluates whether the intended changes, such as modifications in focal subjects or camera configurations, are correctly realized. For each transition, PTCS is computed as the geometric mean of PCP and PTA and then averaged over the evaluation set. By jointly evaluating unchanged elements and intended changes, PTCS provides a more direct and faithful assessment of inter-shot consistency.

\begin{table}[t]
    \centering
    \scriptsize
    \caption{
    \textbf{Theoretical efficiency comparison.}
    FLOPs are calculated for self-attention and retrieval-related operations.
    Collected K/V denotes the logical amount of retrieved memory elements.
    }
    \label{tab:efficiency-theory}
    \small

    \begin{tabular}{lccc}
        \toprule
        Method 
        & Retrieval Overhead 
        & Attention 
        & Total FLOPs \\
        & (TFLOPs)
        & (TFLOPs)
        & (TFLOPs)
        \\
        \midrule

        Full history
        & \textbf{0}
        & 932.56
        & 932.56
        \\

        Frame retrieval ($K=20$)
        & 0.0453
        & 649.97
        & 650.02
        \\

        Cluster retrieval ($K=28$, $S=1024$)
        & 15.0180
        & \textbf{623.80}
        & \textbf{638.82}
        \\

        \bottomrule
    \end{tabular}
\end{table}

\begin{table*}[t]
    \centering
    \scriptsize
    \caption{
    \textbf{Quantitative comparison} on streaming multi-shot video generation.
    The best and second-best results are highlighted in \textbf{bold} and
    \underline{underlined}, respectively.
    }
    \label{tab:results}

    \renewcommand{\arraystretch}{1.15}
    \setlength{\tabcolsep}{4.5pt}

    \begin{tabular}{l@{\hspace{4pt}}c|ccc|cc|cccc}
        \toprule

        \multirow{2}{*}{\textbf{Method}}
        & \multirow{2}{*}{\makecell{\textbf{Params.}}}
        & \multicolumn{3}{c|}{\textbf{Video-level}}
        & \multicolumn{2}{c|}{\textbf{Intra-shot}}
        & \multicolumn{4}{c}{\textbf{Inter-shot}} \\

        \cmidrule(lr){3-5}
        \cmidrule(lr){6-7}
        \cmidrule(lr){8-11}

        &
        &
        \makecell{Trans.\\Ctrl. $\uparrow$}
        &
        \makecell{Text-Video\\Sim. $\uparrow$}
        &
        \makecell{Aesthetic\\$\uparrow$}
        &
        \makecell{Subject\\Cons. $\uparrow$}
        &
        \makecell{Background\\Cons. $\uparrow$}
        &
        \makecell{Inter-shot\\Sim. $\uparrow$}
        &
        PCP $\uparrow$
        &
        PTA $\uparrow$
        &
        \textbf{PTCS} $\uparrow$
        \\

        \midrule

        \multicolumn{11}{l}{\textit{Causal multi-shot models}} \\
        LongLive2 & 5B
        & 0.3856 & 0.2118 & 0.5571 & 0.9496 & 0.9454
        & \underline{0.9420} & \textbf{0.9808} & 0.3764 & 0.3905 \\

        ShotStream & 1.3B
        & 0.9080 & 0.2207 & 0.5776 & \underline{0.9571} & \underline{0.9616}
        & 0.7409 & 0.9501 & 0.6756 & 0.6917 \\

        \midrule

        \multicolumn{11}{l}{\textit{Causal multi-prompt models}} \\
        Infinity-RoPE & 1.3B
        & 0.0124 & 0.2178 & 0.5747 & 0.9159 & 0.9167
        & 0.9396 & 0.9801 & 0.3619 & 0.3881 \\

        Self-Forcing & 1.3B
        & 0.1866 & 0.2208 & 0.5732 & 0.9183 & 0.9152
        & 0.9021 & 0.9609 & 0.2779 & 0.2964 \\

        MemFlow & 1.3B
        & 0.0440 & 0.2163 & \textbf{0.5799} & 0.9456 & 0.9374
        & \textbf{0.9545} & \underline{0.9807} & 0.2143 & 0.2254 \\

        \midrule

        \multicolumn{11}{l}{\textit{Bidirectional models}} \\
        MultiShotMaster & 1.3B
        & \underline{0.9797} & 0.2292 & 0.5701 & 0.9413 & 0.9398
        & 0.6626 & 0.9031 & \underline{0.7956} & \underline{0.7984} \\

        HoloCine & 14B
        & 0.9671 & \underline{0.2315} & 0.5473 & 0.9294 & 0.9360
        & 0.7214 & 0.9250 & 0.7309 & 0.7385 \\

        \midrule

        \methodname{} (\textbf{Ours}) & 1.3B
        & \cellcolor{blue!10}\textbf{0.9960} & \cellcolor{blue!10}\textbf{0.2336} & \cellcolor{blue!10}\underline{0.5779}
        & \cellcolor{yellow!20}\textbf{0.9716} & \cellcolor{yellow!20}\textbf{0.9630}
        & 0.7598 & 0.9270 & \textbf{0.8004} & \cellcolor{green!10}\textbf{0.8242} \\

        \bottomrule
    \end{tabular}
\end{table*}

\begin{table*}[t]
    \centering
    \scriptsize
    \caption{
        \textbf{Ablation study.} The best and runner-up are in \textbf{bold} and \underline{underlined}.
        C: clustered KV cache; F: frame grouping; Q-wise: query- and head-wise retrieval; Shared: shared across queries.
    }
    \label{tab:ablation}

    \renewcommand{\arraystretch}{1.12}
    \setlength{\tabcolsep}{2.0pt}

    \begin{tabular}{@{}lccccc@{\hspace{3pt}}lccccc@{}}
        \toprule

        \textbf{Method}
        & \textbf{T-Ctrl} $\uparrow$
        & \textbf{T-V Sim.} $\uparrow$
        & \textbf{Aes.} $\uparrow$
        & \textbf{Intra.} $\uparrow$
        & \textbf{PTCS} $\uparrow$
        &
        \textbf{Method}
        & \textbf{T-Ctrl} $\uparrow$
        & \textbf{T-V Sim.} $\uparrow$
        & \textbf{Aes.} $\uparrow$
        & \textbf{Intra.} $\uparrow$
        & \textbf{PTCS} $\uparrow$
        \\

        \midrule

        \multicolumn{6}{c}{\textit{Training Strategy}}
        &
        \multicolumn{6}{c}{\textit{Retrieval Strategy (Plug-in)}}
        \\

        \cmidrule(lr){1-6}
        \cmidrule(lr){7-12}

        Teacher Forcing
        & 0.9789 & 0.2306 & 0.5696 & 0.9615 & 0.7120
        &
        Full-memory
        & \textbf{0.9983} & 0.2337 & \textbf{0.5776} & 0.9640 & \underline{0.7922}
        \\

        Resampled Context 
        & \cellcolor{blue!10}\textbf{0.9983} & \cellcolor{blue!10}\textbf{0.2337} & \cellcolor{blue!10}\textbf{0.5776} & \cellcolor{yellow!20}\textbf{0.9640} & \cellcolor{green!10}\textbf{0.7922}
        &
        F + Shared
        & 0.9949 & \underline{0.2379} & 0.5746 & 0.9643 & 0.7280
        \\

        \cmidrule(lr){1-6}
        \multicolumn{6}{c}{\textit{Query-level Retrieval Variant}}
        &
        F + Q-wise
        & 0.9946 & 0.2368 & 0.5755 & \underline{0.9645} & 0.7525
        \\

        \cmidrule(lr){1-6}

        Query-group
        & 0.9953 & \textbf{0.2362} & 0.5748 & 0.9656 & 0.8233
        &
        C + Shared
        & 0.9950 & 0.2378 & 0.5732 & 0.9626 & 0.7442
        \\

        Query-wise (\textbf{Ours})
        & \cellcolor{blue!10}\textbf{0.9960} & \cellcolor{blue!10}0.2336 & \cellcolor{blue!10}\textbf{0.5779} & \cellcolor{yellow!20}\textbf{0.9673} & \cellcolor{green!10}\textbf{0.8242}
        & C + Q-wise (\textbf{Ours})
        & \cellcolor{blue!10}\underline{0.9953} & \cellcolor{blue!10}\textbf{0.2382} & \cellcolor{blue!10}\underline{0.5768} & \cellcolor{yellow!20}\textbf{0.9654} & \cellcolor{green!10}\textbf{0.8007}
        \\

        \bottomrule
    \end{tabular}
\end{table*}

\textbf{Qualitative Results.} Figure \ref{fig:results} presents visual comparisons between \methodname{} and representative causal and bidirectional baselines. Causal multi-prompt models~\citep{yesiltepe2026infinity,ji2025memflow,huang2026self}, despite using varying prompts to enrich the generated content, often struggle to produce explicit shot transitions under multi-shot conditioning and may introduce visual artifacts. Among causal multi-shot baselines, LongLive2~\citep{chen2026longlive2} still exhibits limited transition control, while ShotStream~\citep{luo2026shotstream} tends to lose consistency as the number of shots increases. 
Notably, we do not include CausalCine \citep{meng2026causalcine} in the comparison since its model weights are not publicly available. Instead, we implement a comparable frame-grouping strategy with shared retrieval, which is further analyzed in our ablation study.
Strong bidirectional baselines~\citep{wang2026multishotmaster,meng2026holocine} also occasionally suffer from cross-shot inconsistency or incorrect shot counts. In contrast, \methodname{} generates stable multi-shot videos in an autoregressive setting while faithfully following the prescribed transitions and maintaining cross-shot consistency, including challenging revisitation scenarios.

\textbf{Quantitative Results.} Table \ref{tab:results} reports the quantitative comparison on our evaluation set. \methodname{} achieves strong performance across video-level quality and intra-shot consistency metrics, demonstrating its ability to generate high-quality videos while maintaining stable shot-level content. For inter-shot evaluation, \methodname{} achieves the best performance on PTCS, which jointly considers unchanged elements and intended changes during shot transitions. Notably, models with high similarity-based inter-shot scores or PCP values often exhibit limited transition control, while models specifically designed for multi-shot generation may still receive relatively low scores on these metrics. This indicates that similarity-based or preservation-only metrics can overestimate inter-shot consistency when the intended transition is not properly realized. By explicitly considering both preserved and changing elements specified in the prompt, PTCS provides a more faithful evaluation of transition quality. Finally, \methodname{} achieves superior PTCS performance and even surpasses larger bidirectional models with substantially more parameters.

\subsection{Ablation Study}
\label{sec:ablation}

To validate the effectiveness of different design choices in \methodname{}, we conduct ablation studies from three perspectives, with results summarized in Table \ref{tab:ablation}.

\textbf{Training Strategy.} In Section \ref{sec:training}, we introduce resampled history context to simulate autoregressive error accumulation during training. Compared with the standard teacher-forcing strategy, resampling context training provides more consistent training conditions with autoregressive inference and achieves superior performance across different metrics, especially in long-horizon consistency.

\textbf{Retrieval Strategy.} We analyze different retrieval strategies introduced in Sections \ref{sec:clusterKV} and \ref{sec:query} under a plug-in setting. Specifically, we compare our clustered KV cache and query-wise retrieval with frame-level grouping and shared retrieval. To isolate the effect of retrieval design, we directly apply different strategies to the adapted base model without additional training. As shown, query-wise retrieval consistently improves performance under both frame-level and clustered memory organization, demonstrating the importance of fine-grained memory access. Moreover, clustered KV cache further improves transition quality and inter-shot consistency over frame-level grouping, validating the effectiveness of semantic-based memory organization. Notably, without additional training, our retrieval strategy achieves comparable performance to full-memory attention while obtaining higher inter-shot consistency, demonstrating its effectiveness under a constrained memory budget.

\textbf{Query-level Retrieval Variant.} We investigate a query grouping variant of our retrieval mechanism. Instead of performing independent retrieval for each query, this variant recursively partitions queries into groups according to their spatial and semantic similarity using an adaptive tree structure and shares retrieved memory within each group.
Although query grouping provides an effective approximation by exploiting query similarity, it remains slightly inferior to fully query-wise retrieval, further validating the importance of fine-grained query retrieval.

%% file: sec/5_conclusion.tex
\section{Conclusion}

In this paper, we introduce \methodname{}, a fine-grained memory retrieval framework for autoregressive multi-shot video generation. Combining an online clustered KV cache with query-adaptive retrieval, \methodname{} enables broader historical access and improves long-range consistency in revisitation scenarios. We develop an autoregressive adaptation framework with resampled context to analyze intrinsic memory-access patterns, revealing semantic clustering among memory tokens and heterogeneous query-head preferences that motivate our design. Extensive experiments demonstrate the effectiveness of our framework.
Our results highlight the importance of fine-grained memory modeling for autoregressive video generation. Rather than relying only on recent frames, autoregressive models should exploit information throughout history. Advances in memory organization and retrieval will enable scalable long-horizon video generation.

%% file: sec/A_memory_access.tex
\section{Details of Memory-Access Pattern}
\label{sec:sup_ana}

In this section, we provide further analysis of the memory selection patterns discussed in Section \ref{sec:insight}. Figure \ref{fig:query} presents the heatmaps of query-head memory preferences and the resulting memory coverage achieved by selecting different percentages of memory tokens for each query-head pair. The query-head pairs shown in the figure are randomly sampled for visualization. In Section \ref{sec:sup_heterogeneous}, we provide a more comprehensive analysis to demonstrate the heterogeneous memory selection behaviors across queries and attention heads. Section \ref{sec:pre-post} investigates the clustering behaviors before and after positional encoding, providing further evidence for our memory organization design. Finally, Section \ref{sec:sup_hit} analyzes the retrieval overlap between \methodname{} and full-memory attention, demonstrating that our proposed memory access mechanism can effectively approximate full-memory access patterns during autoregressive rollout.

\subsection{Heterogeneous Query and Head Memory Preferences}
\label{sec:sup_heterogeneous}

In Figure \ref{fig:query}, we provide an overall visualization of the heterogeneous memory selection behaviors across queries and attention heads. The memory coverage curve shows that when each query-head pair selects only the top 30\% memory tokens with the highest attention probabilities, the aggregated receptive field can cover nearly all historical memory frames. In the heatmap, each row corresponds to a query-head pair, while each column represents a spatial memory unit from a historical frame. Specifically, each frame is divided into $2\times2$ spatial regions, resulting in four memory units per frame. The color intensity indicates the ratio of selected tokens within each memory unit among the top 30\% tokens selected by the corresponding query-head pair, reflecting its preference over different historical regions. Due to the large number of query-head pairs, only a subset of rows is randomly sampled for visualization while preserving all historical memory units, providing an overview of the selection patterns.

Although the differences among highlighted regions across rows in Figure \ref{fig:query} already reveal the existence of heterogeneous memory selection over the entire history sequence, the diversity of query preferences is also manifested within individual memory frames. However, the $2\times2$ spatial partition provides only a coarse characterization of such intra-frame variations. To further analyze fine-grained spatial selection behaviors, we focus on two representative memory frames, adopt a finer-grained memory unit partition, and provide additional visualizations of query-specific and head-specific memory preferences.

\begin{figure} 
    \begin{center}
        \includegraphics[width=0.9\linewidth]{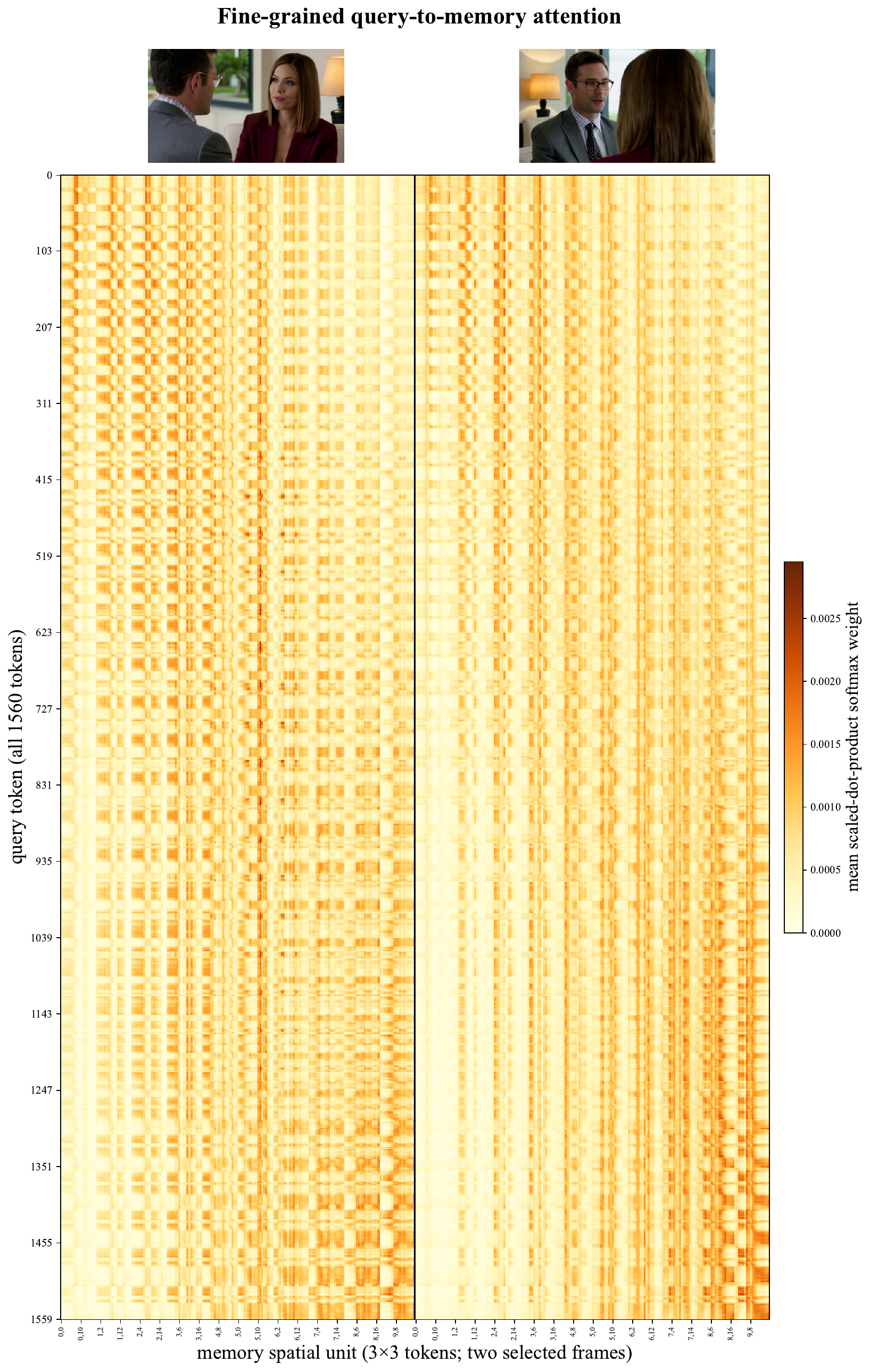}
    \end{center}
    \caption{\textbf{Visualization of query-wise memory selection diversity.} For the selected memory frames, different queries exhibit distinct spatial preferences within the same frame, revealing heterogeneous memory selection behaviors. Each query corresponds to the first latent of the current chunk, and each memory unit is formed by grouping neighboring \(3\times3\) spatial tokens.} 
   \label{fig:sup_query}  
\end{figure}

We first focus on the heterogeneous memory selection behaviors across different queries. In Figure \ref{fig:sup_query}, we analyze how query tokens from the current chunk attend to local regions within historical memory frames. We select representative memory frames from different shots and visualize the spatial preferences of different queries. In contrast to Figure \ref{fig:query}, where each frame is aggregated into four coarse spatial regions using a $2\times2$ partition, here we adopt a finer token-level visualization by grouping neighboring $3\times3$ spatial tokens into individual memory units.
The visualization reveals clear differences in the memory preferences of different queries, even within the same historical frame. This indicates that heterogeneous memory selection is not only manifested across different frames in the entire memory sequence, but also exists among local spatial regions within a single frame. These observations further demonstrate the necessity of fine-grained memory retrieval.

\begin{figure} 
    \begin{center}
        \includegraphics[width=0.9\linewidth]{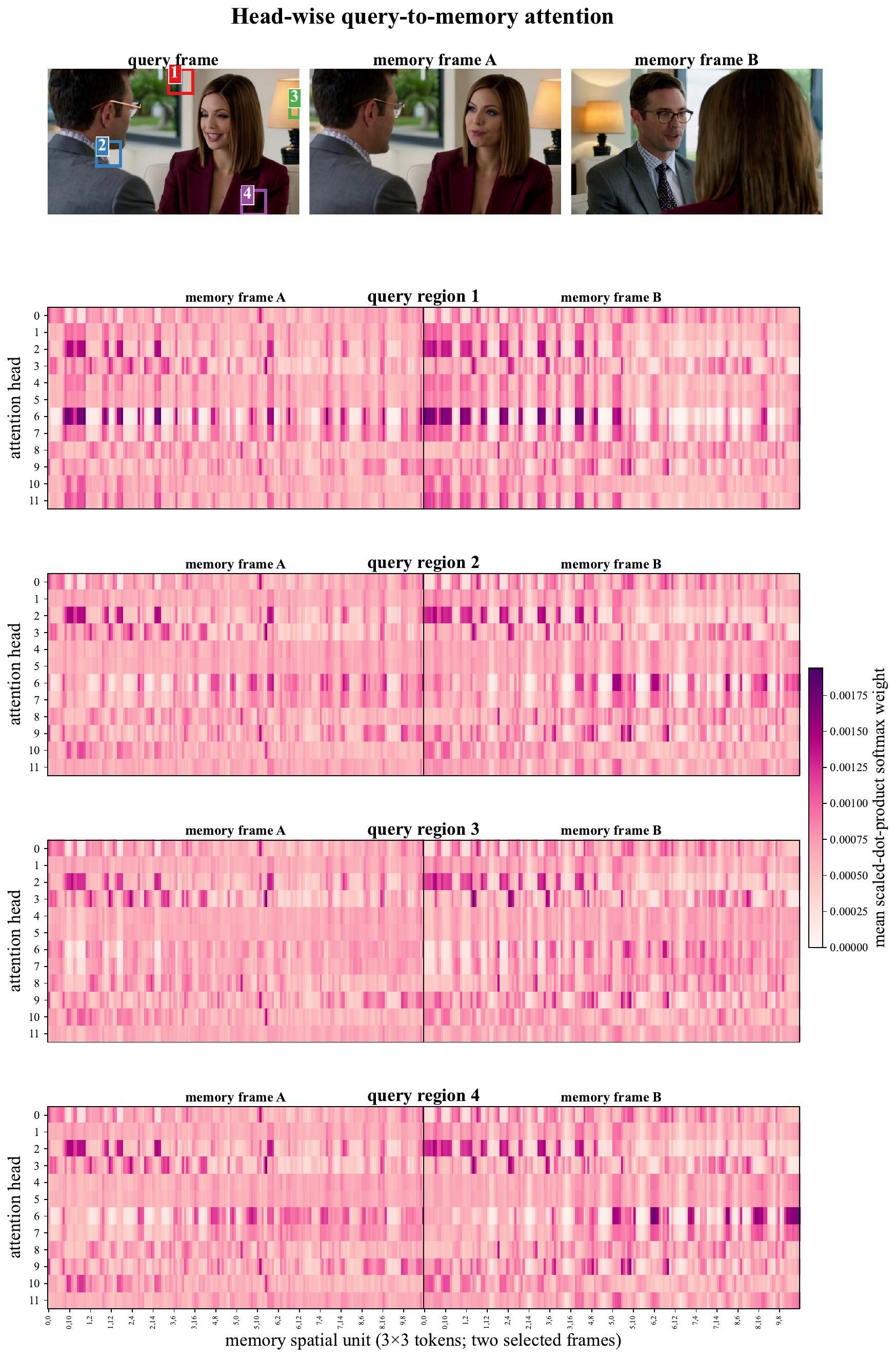}
    \end{center}
    \caption{\textbf{Visualization of head-wise memory selection diversity.} For the selected memory frames, different attention heads exhibit distinct memory preferences. We visualize multiple query regions for generality. The variations among highlighted regions across rows in each heatmap illustrate the diversity of head-wise memory selection.} 
   \label{fig:sup_head}  
\end{figure}

To further investigate the diversity of memory selection across attention heads, we conduct the same analysis from the head-wise perspective and visualize the results in Figure \ref{fig:sup_head}. For different attention heads, we show their attention probabilities over two selected memory frames and observe substantial variations in their selection patterns. Consistent with the motivation of multi-head attention \citep{vaswani2017attention}, different heads focus on different types of information and thus play distinct roles in memory selection. These observations further support the necessity of independent retrieval across attention heads.

\begin{figure}[!t]
    \begin{center}
        \includegraphics[width=1.0\linewidth]{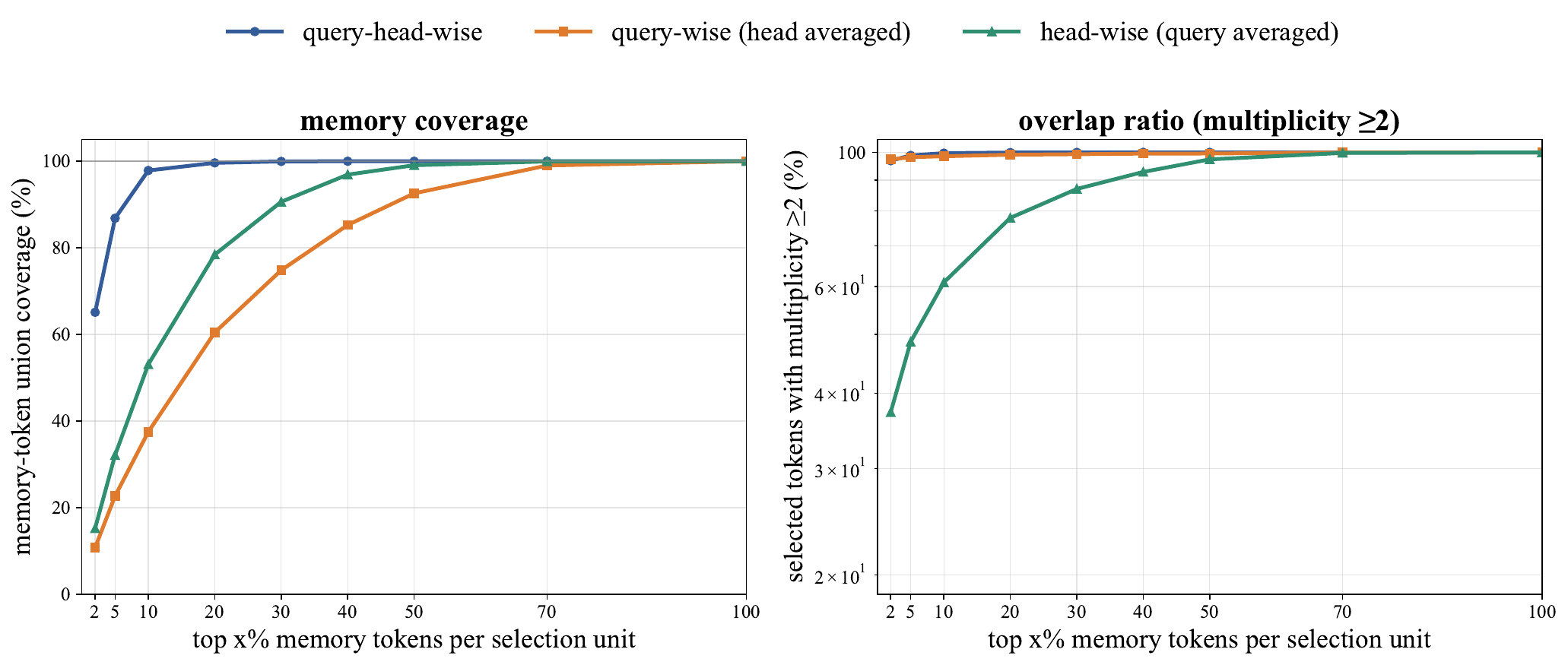}
    \end{center}
    \caption{Comparison of \textbf{memory-token coverage and selection overlap} under three selection granularities: query-head-wise, query-wise with head averaging, and head-wise with query averaging. The horizontal axis denotes the fraction of memory tokens selected by each selection unit. The left panel reports the percentage of distinct memory-token positions covered by the union of all selections, while the right panel shows the percentage of selected positions with multiplicity at least two.} 
   \label{fig:curve}  
\end{figure}

Finally, the lower-right inset of Figure \ref{fig:query} provides the memory coverage curve for query-head-wise retrieval. To further characterize the heterogeneous preferences of queries and attention heads, we analyze memory coverage and retrieval overlap ratio from the query-wise and head-wise perspectives, as shown in Figure \ref{fig:curve}. We observe that query-head-wise retrieval achieves the largest memory coverage, while separating query-wise and head-wise selection reveals stronger diversity among attention heads. Specifically, head-wise retrieval exhibits faster coverage growth and lower overlap ratios compared with query-wise retrieval, indicating more diverse memory preferences across attention heads and further motivating independent head-wise retrieval.

In general, these analyses provide further evidence for heterogeneous query and head memory preferences, which motivate the design of query-wise and head-wise retrieval in \methodname{}.

\subsection{Pre-RoPE Clustering vs. Post-RoPE Clustering}
\label{sec:pre-post}

\begin{figure}[!t]
    \begin{center}
        \includegraphics[width=1.0\linewidth]{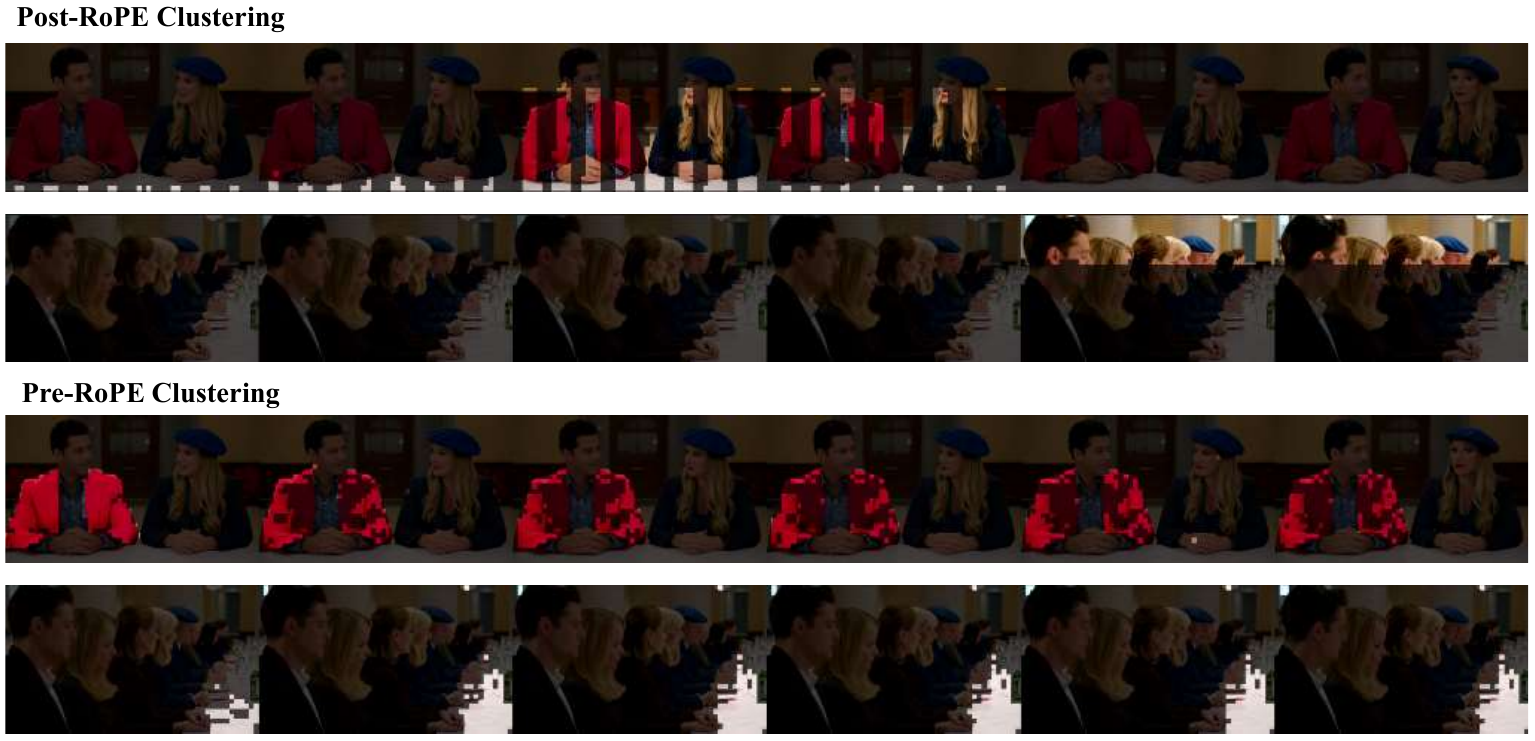}
    \end{center}
    \caption{\textbf{Visualization of token clustering before and after RoPE.} Compared with Pre-RoPE representations, Post-RoPE similarities exhibit stronger positional dependency and weaker semantic clustering structures.} 
   \label{fig:pre-post}  
\end{figure}

In Section \ref{sec:clusterKV}, we introduce an online clustering mechanism to organize memory tokens into semantic groups rather than predefined frame-based partitions. In practice, before entering each transformer block, token representations are augmented with positional encoding. Our clustering and retrieval operations are therefore performed before RoPE (Pre-RoPE). This design choice is motivated by the observation that positional encoding introduces strong absolute position information into token representations, which can obscure semantic similarity and weaken semantic clustering structures.

In fact, the visualization in Figure \ref{fig:memory} is obtained using token representations before positional encoding. To further investigate this effect, Figure \ref{fig:pre-post} presents representative clustering results before and after applying RoPE. We observe that RoPE-transformed tokens exhibit stronger position-dependent similarity patterns, while the semantic grouping structure becomes less apparent. Therefore, performing clustering and retrieval after positional encoding may bias memory organization toward positional grouping and degrade into a behavior similar to frame-level partitioning, limiting the benefits of semantic-aware memory organization. These observations support our choice of conducting clustering and retrieval in the Pre-RoPE space.

\subsection{Retrieval Alignment with Full-Memory Attention}
\label{sec:sup_hit}

\begin{figure}[!t]
    \begin{center}
        \includegraphics[width=1.0\linewidth]{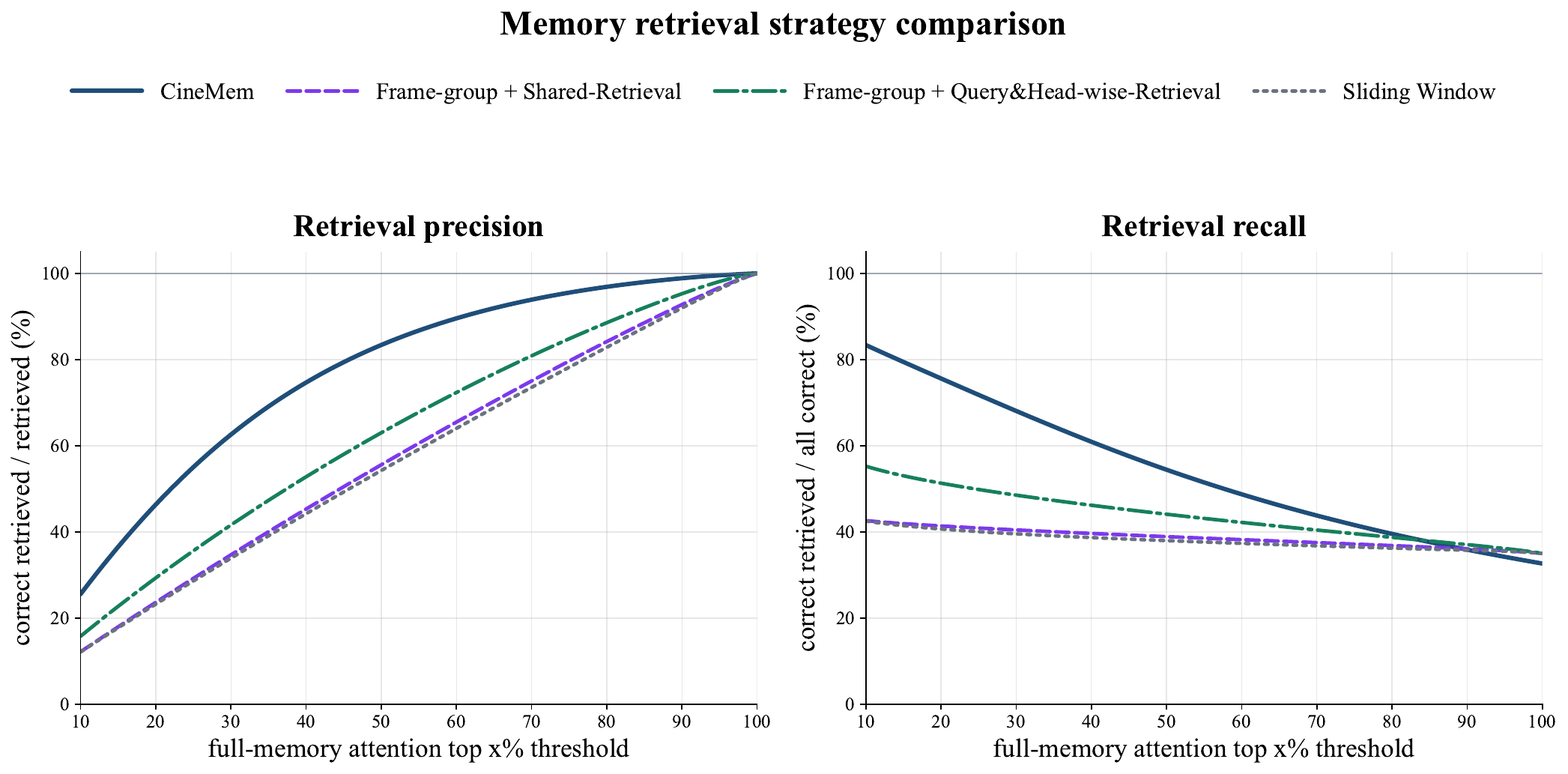}
    \end{center}
    \caption{Precision (left) and recall (right) of different retrieval strategies under varying full-memory attention thresholds. For each query-head pair, full-memory attention serves as the oracle, and retrieved tokens are considered matched if they fall within the top-\(x\%\) oracle-ranked memory tokens. Online-clustered KV cache retrieves tokens from query-specific selected clusters, while Latent-group, Frame-group, and Sliding Window perform frame-level or fixed-context retrieval. All curves are averaged over attention blocks. Online-clustered KV cache achieves a better precision-recall trade-off, providing a selective approximation to full-memory access.} 
   \label{fig:sup_hit}  
\end{figure}

Our retrieval mechanism aims to approximate the memory access pattern of full-memory attention under a constrained retrieval budget. To further evaluate this property, we analyze retrieval alignment from the perspective of full-memory attention. Specifically, we treat full-memory attention as an oracle: for each query-head pair, memory tokens are ranked according to their attention scores, and highly attended tokens are regarded as the oracle retrieval targets.

We compare different retrieval strategies under varying selection thresholds. A retrieved token is considered matched if it belongs to the top-\(x\%\) memory tokens ranked by the full-memory attention distribution, where the ranking is determined by the corresponding query-key similarity using normalized dot products. We report precision, defined as the proportion of retrieved tokens overlapping with the oracle top-\(x\%\) set, and recall, defined as the proportion of oracle top-\(x\%\) tokens successfully recovered by the retrieval strategy:
\begin{equation}
    \mathcal{O}_{q,h}^{x} = \mathrm{Top}_{x\%}
    \left(A^{\mathrm{full}}_{q,h}\right),
    \quad
    P_x =
    \frac{
        |\mathcal{R}_{q,h}\cap \mathcal{O}_{q,h}^{x}|
    }{
        |\mathcal{R}_{q,h}|
    },
    \quad
    R_x =
    \frac{
        |\mathcal{R}_{q,h}\cap \mathcal{O}_{q,h}^{x}|
    }{
        |\mathcal{O}_{q,h}^{x}|
    } ,
    \label{eq:retrieval_alignment}
\end{equation}
where \(A^{\mathrm{full}}_{q,h}\) denotes the full-memory attention distribution of query-head pair \((q,h)\), \(\mathcal{O}_{q,h}^{x}\) represents the oracle top-\(x\%\) memory tokens, and \(\mathcal{R}_{q,h}\) denotes the tokens retrieved by a given strategy.
Both metrics are computed independently for each query-head pair and then averaged across query-head pairs and attention blocks.

As shown in Figure \ref{fig:sup_hit}, \methodname{} achieves consistently higher precision than frame-level grouping and sliding-window retrieval across almost the entire threshold range. It also achieves higher recall in the low-to-moderate threshold regime, indicating that it can recover a larger fraction of highly relevant memory tokens while introducing fewer irrelevant tokens. At very high thresholds, the recall of the online-clustered KV cache becomes slightly lower than fixed-budget baselines. This behavior does not indicate inferior retrieval quality. When \(x=100\%\), the oracle set contains all memory tokens, and recall mainly reflects the proportion of memory tokens retrieved. Fixed-budget baselines always retrieve a predetermined number of tokens, whereas the online-clustered KV cache adaptively retrieves a compact set of relevant clusters. Therefore, the reduced recall at large \(x\) reflects a smaller and less redundant retrieval set rather than a failure to identify important memory information.

Overall, these results demonstrate that \methodname{} provides a more selective approximation to full-memory attention, achieving a better precision-recall trade-off while preserving access to highly relevant historical information.

%% file: sec/B_Methodology.tex
\section{Details of Methodology}
\label{sec:sup_retrieval}

In this section, we provide additional details of the retrieval mechanisms introduced in Section~\ref{sec:clusterKV} and Section \ref{sec:query}, including the implementation of the clustered KV cache and query-adaptive retrieval. We further describe the post-training procedure for adapting the model with the proposed retrieval framework.

\subsection{Online Clustered KV Cache}
\label{sec:sup_online_KV}

We provide additional implementation details of the online-clustered KV cache introduced in Section \ref{sec:clusterKV}. The cache is maintained independently for each transformer block, attention head, and classifier-free guidance branch. For each attention head $h$, the clustered historical memory is represented as:
\begin{equation}
    \mathcal{C}_h =
    \left\{
    (k_i,v_i,z_i)
    \right\}_{i=1}^{T},
\end{equation}
where \(k_i\) and \(v_i\) denote the key and value of the \(i\)-th historical memory token, and \(z_i\) denotes its cluster assignment. Each active cluster \(c\) maintains its token count \(n_c\), accumulated key vector \(s_c\), and normalized centroid:
\begin{equation}
    \mu_c = \frac{s_c}{\|s_c\|_2}.
\end{equation}

The KV tensors are stored separately from cluster metadata. Specifically, pre-RoPE keys are used for semantic clustering and query routing, while the corresponding post-RoPE keys are retained for final attention computation.

The complete online update procedure is summarized in Algorithm~\ref{alg:online_clustered_kv}. Below, we provide additional details of each step.

\begin{algorithm}[H]
\caption{Online Update Procedure of the Head-wise Clustered KV Cache}
\label{alg:online_clustered_kv}
\begin{algorithmic}[1]
\REQUIRE Existing cache $\mathcal{C}_h$,
newly generated key-value chunk $(K_{\mathrm{new}},V_{\mathrm{new}})$,
target cluster size $S$, and maximum cluster size $S_{\max}$
\ENSURE Updated clustered cache $\mathcal{C}_h$ and routing metadata

\STATE If adjacent-frame retention is enabled, move the newest latent frame to the recent-memory buffer and append the remaining keys and values to the clusterable cache
\STATE Normalize the clusterable keys independently for head $h$ to obtain
$\hat{K}$
\IF{$\mathcal{C}_h$ is empty}
    \STATE Set
    $G_0 \leftarrow \max(1,\lceil |\hat{K}|/S\rceil)$
    \STATE Select $G^\star$ diverse seeds from $\hat{K}$ using farthest-point
    sampling
    \STATE Assign every token to its most similar seed and run several Lloyd
    refinement iterations
\ELSE
    \STATE Append the new keys and values to $\mathcal{C}_h$
    \STATE Compute
    $G^\star \leftarrow \lceil \left|\mathcal{C}_h\right|_{\text{token}}/S\rceil$
    \STATE Activate additional cluster slots when the current number of
    clusters is smaller than $G^\star$
    \STATE Seed new clusters with low-similarity new tokens using
    farthest-point sampling
    \STATE Assign new tokens according to Equation~\ref{eq:cluster}, using the
    load penalty and assignment-consistency term
    \STATE Recompute cluster sums, counts, centroids, and assignments over
    all tokens in $\mathcal{C}_h$
    \STATE Run global Lloyd refinement for a fixed number of iterations
\ENDIF

\FOR{each split pass}
    \IF{there exists a cluster $c$ with $n_c>S_{\max}$}
        \STATE Select the least similar token in $c$ as a child-cluster seed
        \STATE Partition the members of $c$ into two balanced subsets
        \STATE Recompute statistics and refine the assignments
    \ENDIF
\ENDFOR

\FOR{each merge pass}
    \IF{the number of active clusters exceeds $G^\star$}
        \STATE Select the smallest cluster as a donor
        \STATE Find the most similar compatible target cluster whose merged
        size does not exceed $S_{\max}$
        \STATE Merge the donor into the target and recompute cluster statistics
    \ENDIF
\ENDFOR

\STATE Reorder the historical keys and values by cluster assignment independently
for each head
\STATE Compute cumulative cluster offsets and compact active-cluster metadata
\RETURN $\mathcal{C}_h$
\end{algorithmic}
\end{algorithm}

\textbf{Initialization.} 
At the beginning of rollout, the cache is empty. When the first generated chunk becomes available, its keys are normalized independently for each attention head and initialized into clusters. Given \(T_0\) available memory tokens and target cluster size \(S\), the initial number of clusters is:
\begin{equation}
    G_0 = \max\left(1,\left\lceil \frac{T_0}{S}\right\rceil\right).
\end{equation}
In our implementation, \(S=1024\). Initial centroids are selected using farthest-point sampling to provide diverse seeds in the key space, followed by several Lloyd refinement iterations.

\textbf{Online Cluster Assignment.}
After each generated chunk, newly generated key-value pairs are appended to the persistent cache. When adjacent-frame retention is enabled, the newest latent frame is temporarily maintained in a separate recent-memory buffer. During the next update, the previous recent frame is incorporated into clusterable memory, while the newest frame remains directly accessible. This implements the latest-memory anchor in Equation~\ref{eq:query}. The initial chunk is never removed from the historical cache and remains available throughout the rollout.

For each newly appended token, cluster assignment is determined by the cosine similarity between its normalized key and active cluster centroids. The assignment rule in Equation~\ref{eq:cluster} is implemented as:
\begin{equation}
z_i =
\arg\max_c
\left(
\hat{k}_i^{\top}\mu_c
-\lambda
\operatorname{clip}\left(
\max\left(\frac{n_c}{S}-1,0\right),0,1
\right)
+\eta\mathbb{I}[c=z_i^{\mathrm{old}}]
\right),
\end{equation}
where \(\lambda=0.02\) and \(\eta=0.02\). The load penalty discourages assignments to clusters exceeding the target size, while the consistency term stabilizes assignments during online refinement.

\textbf{Dynamic Cluster Maintenance.} 
As autoregressive rollout proceeds, the number of historical memory tokens continuously increases. Maintaining a fixed number of clusters would enlarge the average cluster size and reduce the semantic coherence of cluster-level retrieval. Therefore, we dynamically adjust the number of active clusters according to the target cluster size \(S\).

Given $T$ clusterable historical tokens, the desired number of active clusters is
\begin{equation}
    G^\star =
  \left\lceil \frac{T}{S}\right\rceil.
\end{equation}
Here, \(G^\star\) represents a target cluster count rather than a strict upper bound. The number of active clusters may temporarily exceed this value after split operations and is reduced through subsequent capacity-aware merging.

When newly generated tokens arrive, additional clusters are activated only when the current number of clusters is smaller than \(G^\star\). New centroids are initialized using farthest-point sampling from tokens with low similarity to existing centroids. The number of newly activated clusters is constrained by available cluster slots, the number of new tokens, and the per-update activation limit. After assignment, cluster statistics are recomputed over all historical tokens, and global refinement is performed, allowing previous tokens to change assignments when the centroid configuration changes.

\textbf{Cluster Balancing and Retrieval Preparation.}
To avoid highly imbalanced clusters, we further apply explicit split and merge operations as shown in Figure~\ref{fig:method}. A cluster is considered over-sized when its occupancy exceeds
$S_{\max}=1200$. Such a cluster is split by selecting its least similar token as a new seed and partitioning the original members into two approximately balanced subsets, followed by additional Lloyd refinement. If the number of active clusters is larger than $G^\star$, the smallest compatible cluster is merged into the most similar neighboring cluster, provided that the resulting
cluster does not exceed $S_{\max}$. The complete online update  procedure is summarized in Algorithm~\ref{alg:online_clustered_kv}.

To avoid highly imbalanced clusters, we apply explicit split and merge operations as illustrated in Figure~\ref{fig:method}. A cluster is considered oversized when its occupancy exceeds \(S_{\max}=1200\). Such clusters are split by selecting the least similar token as a new seed and partitioning the cluster members into two approximately balanced subsets, followed by additional Lloyd refinement. When the number of active clusters exceeds \(G^\star\), the smallest compatible cluster is merged into its most similar neighboring cluster, provided that the resulting cluster size does not exceed \(S_{\max}\).

After the update, tokens are reordered according to their cluster labels independently for each head, making tokens within the same cluster contiguous in memory. The implementation additionally maintains cumulative offsets, compact centroid tensors, active-cluster masks, and mappings between compact cluster indices and original cluster slots. These structures allow retrieval to gather complete clusters efficiently without scanning the entire historical KV cache.

\subsection{Query-adaptive Retrieval}
\label{sec:sup_query}

The query-adaptive retrieval module is implemented within each attention layer. Let the current chunk contain $L$ query tokens. The projected query tensor has shape $[B,L,H,D]$, where $H$ denotes the number of attention heads. For each query position $\ell$ and head $h$, the normalized query is compared with the compact centroids of the corresponding clustered cache:
\begin{equation}
    r_{\ell,h,c} =
\operatorname{Normalize}(q_{\ell,h})^{\top}\mu_{h,c}.
\end{equation}

The active clusters with the highest scores are selected independently for each query-head pair:
\begin{equation}
    \mathcal{S}_{q,h}
=
\operatorname{TopK}_{c}
\left(
r_{\ell,h,c}
\right).
\end{equation}
This implements the retrieval operation defined in Equation~\ref{eq:query}.

The routing decision is performed using pre-RoPE queries and normalized cluster centroids. The selected clusters are then used to gather the corresponding historical keys and values from the reordered cache. The retrieved pre-RoPE keys are transformed with positional encoding to obtain post-RoPE keys for final attention computation. Therefore, semantic routing is performed in a position-independent representation space, while positional information is preserved during attention calculation.

If the total number of historical tokens is smaller than the retrieval budget, selective routing is bypassed and full historical attention is applied. Otherwise, the selected cluster indices form a routing tensor of shape $[B,H,L,K_{\mathrm{ret}}]$. The retrieved memory corresponds to Equation~\ref{eq:query}, where the first and latest terms represent the retained boundary context, and the middle term is assembled from clusters selected specifically for each query-head pair.

For computational efficiency, attention computation is divided into two branches. The first branch performs attention over the current chunk and the directly retained recent memory. The second branch applies FlashAttention to the historical tokens gathered from the selected clusters. The selected-cluster outputs are combined using their log-sum-exp normalization and merged with the current-chunk branch using the corresponding attention normalizers. Thus, retrieved memory is integrated within the original attention normalization rather than introduced as an independent residual feature.

The clustered cache is updated only after a generated chunk completes its denoising trajectory. Conditional and unconditional caches are maintained separately for classifier-free guidance, and the same query-head-wise retrieval procedure is applied to both branches. During subsequent denoising steps, each query-head pair can therefore access a different subset of historical clusters, while the current chunk remains fully accessible.

\subsection{Training with Memory Retrieval Mechanism}
\label{sec:sup_train}

In Section \ref{sec:training}, we follow the resampled context strategy \citep{guo2025end} during autoregressive adaptation, where model-generated frames are used as historical context to simulate error accumulation and facilitate long-horizon multi-shot video generation. As discussed in Section \ref{sec:query}, the same error simulation mechanism is further adopted when training with memory retrieval. In this section, we provide additional details of the training procedure with the proposed memory retrieval
framework. The complete procedure is summarized in Algorithm~\ref{alg:training_memory_retrieval}.

\begin{algorithm}[t]
\caption{Training with sequential memory retrieval and resampled context.}
\label{alg:training_memory_retrieval}
\begin{algorithmic}[1]

\REQUIRE Model $v_{\theta}$, clean latent chunks
$\{\mathbf{x}^{0}_{k}\}_{k=1}^{N}$, chunk-wise contexts, and noise scheduler
\ENSURE Training loss and updated model parameters

\STATE Initialize an empty memory cache $\mathcal{C}_{<1}$

\STATE Sample independent training noise levels
$\{\sigma_k\}_{k=1}^{N}$ and noises
$\{\epsilon_k\}_{k=1}^{N}$

\STATE Construct training inputs
$\mathbf{x}^{\mathrm{train}}_{k}$ and flow-matching targets
$\mathbf{v}^{*}_{k}$ for all chunks

\STATE Sample $\sigma_{\mathrm{low}}$ and construct low-noise chunks
$\{\widetilde{\mathbf{x}}_{k}\}_{k=1}^{N}$ using the resampled context strategy

\FOR{$k=1$ to $N$}

    \STATE Select the global and current-shot text contexts for chunk $k$

    \STATE Compute
    $\widehat{\mathbf{v}}^{\mathrm{train}}_{k}
    =
    v_{\theta}
    (\mathbf{x}^{\mathrm{train}}_{k},
    \sigma_k,
    \mathcal{C}_{<k})$
    with query-adaptive memory retrieval

    \STATE Accumulate the chunk loss
    \[
    \mathcal{L}_{k}
    =
    \|
    \widehat{\mathbf{v}}^{\mathrm{train}}_{k}
    -
    \mathbf{v}^{*}_{k}
    \|_2^2
    \]

    \STATE Backpropagate $\mathcal{L}_{k}$ while keeping
    $\mathcal{C}_{<k}$ detached

    \STATE Disable gradients and compute the one-step low-noise prediction
    \[
    \widehat{\mathbf{v}}^{\mathrm{low}}_{k}
    =
    v_{\theta}
    (\widetilde{\mathbf{x}}_{k},
    \sigma_{\mathrm{low}},
    \mathcal{C}_{<k})
    \]

    \STATE Estimate the generated clean chunk
    \[
    \widehat{\mathbf{x}}^{0}_{k}
    =
    \widetilde{\mathbf{x}}_{k}
    -
    \sigma_{\mathrm{low}}
    \widehat{\mathbf{v}}^{\mathrm{low}}_{k}
    \]

    \STATE Run a zero-noise-level forward pass on
    $\widehat{\mathbf{x}}^{0}_{k}$ to extract key-value representations

    \STATE Detach the new key-value tensors, append them to the cache,
    and update online clusters and routing metadata

    \STATE Set
    $\mathcal{C}_{<k+1}
    \leftarrow
    \mathcal{C}_{\leq k}$

\ENDFOR

\STATE Update $\theta$ using the accumulated loss

\RETURN Updated parameters $\theta$

\end{algorithmic}
\end{algorithm}

Training with the proposed memory retrieval mechanism must preserve the causal cache evolution used during inference. In particular, the historical memory available to chunk $k$ should be determined by preceding generated chunks, rather than by a single forward pass over the complete video with an attention mask. We therefore train the model using a sequential chunk-level rollout. The memory cache is initialized as empty and updated only after the current
chunk has been processed. Consequently, the retrieval clusters and
query-head-wise routing used for chunk $k$ are constructed from the same type of progressively accumulated memory as during inference.

Let $\mathbf{x}^{0}_{k}$ denote the clean latent representation of chunk $k$. For the supervised denoising objective, we independently sample a training noise level $\sigma_k$ and a noise tensor $\epsilon_k$ for each chunk. The training input and flow-matching target are defined as:
\begin{equation}
    \mathbf{x}^{\mathrm{train}}_{k}
    =
    (1-\sigma_k)\mathbf{x}^{0}_{k}
    +
    \sigma_k\epsilon_k,
\end{equation}

\begin{equation}
    \mathbf{v}^{\ast}_{k}
    =
    \epsilon_k-\mathbf{x}^{0}_{k}.
\end{equation}

The model is evaluated on $\mathbf{x}^{\mathrm{train}}_{k}$ using the current memory cache $\mathcal{C}_{<k}$, and the denoising loss is computed only for the current chunk:
\begin{equation}
    \mathcal{L}_{k}
    =
    \left\|
    v_{\theta}
    \left(
    \mathbf{x}^{\mathrm{train}}_{k},
    \sigma_k,
    \mathcal{C}_{<k}
    \right)
    -
    \mathbf{v}^{\ast}_{k}
    \right\|_2^2 .
\end{equation}

Importantly, the loss forward pass is performed before the current chunk is inserted into the memory cache. Therefore, the current prediction cannot access its own future key-value representations through the retrieval branch.

The cache update follows the resampled context procedure. We sample one low-noise level $\sigma_{\mathrm{low}}$ for the whole training example using the same shifted sampling strategy as in the adaptation stage:

\begin{equation}
    u\sim\mathcal{N}(0,1),\qquad
    r=\operatorname{sigmoid}(u),\qquad
    \sigma_{\mathrm{low}}
    =
    \frac{\gamma r}{1+(\gamma-1)r},
\end{equation}
where $\gamma$ is the shift parameter. A low-noise version of each chunk is then constructed as:
\begin{equation}
    \widetilde{\mathbf{x}}_{k}
    =
    (1-\sigma_{\mathrm{low}})
    \mathbf{x}^{0}_{k}
    +
    \sigma_{\mathrm{low}}
    \epsilon^{\mathrm{low}}_{k}.
\end{equation}

After computing the supervised loss for chunk $k$, the model performs a one-step low-noise prediction using the same historical cache:
\begin{equation}
    \widehat{\mathbf{v}}_{k}
    =
    v_{\theta}
    \left(
    \widetilde{\mathbf{x}}_{k},
    \sigma_{\mathrm{low}},
    \mathcal{C}_{<k}
    \right),
\end{equation}
\begin{equation}
    \widehat{\mathbf{x}}^{0}_{k}
    =
    \widetilde{\mathbf{x}}_{k}
    -
    \sigma_{\mathrm{low}}
    \widehat{\mathbf{v}}_{k}.
\end{equation}

The estimated clean chunk $\widehat{\mathbf{x}}^{0}_{k}$, rather than the ground-truth chunk $\mathbf{x}^{0}_{k}$, is subsequently forwarded at $t=0$ to extract its key-value representations. These representations are detached from the computation graph and appended to the memory cache. The online clustering, cluster refinement, split/merge balancing, and cache layout construction are then performed using the same procedures as during inference.
The updated cache $\mathcal{C}_{\leq k}$ is used for the next chunk.

This procedure provides two forms of consistency. First, sequential cache rollout exposes the model to the same causal memory organization and query-adaptive retrieval process used during inference. Second, the low-noise one-step prediction introduces accumulated modeling errors into subsequent memory states, following the resampled context strategy used during autoregressive adaptation. The cache itself is treated as a detached state:
gradients are propagated through the current loss forward pass, but not through previous chunks, cluster assignments, or the resampled context rollout. During training, the top-$K$ retrieval routes are computed from detached queries and reused within the corresponding transformer block, while the selected-cluster attention operation remains differentiable with respect to model parameters.

%% file: sec/C_efficiency.tex
\section{Analysis of Efficiency}
\label{sec:efficiency}

In this section, we analyze the efficiency of \methodname{} and compare it
with two representative memory access strategies. Full-history attention is
considered as an upper bound in terms of memory availability, as it provides
access to the complete historical context without retrieval constraints.
Meanwhile, frame-group retrieval serves as an efficient coarse-grained
baseline by selecting a fixed number of historical frames with shared routing.
In contrast, \methodname{} performs fine-grained query-head-wise retrieval over
online-clustered memory, aiming to achieve a better balance between retrieval
granularity and computational cost.
We evaluate these strategies from both theoretical and practical perspectives.
The theoretical analysis focuses on self-attention computation and retrieval
overhead, while the practical evaluation measures actual inference efficiency.

For the theoretical efficiency analysis, we follow the 15-second generation setting of Wan2.1-T2V-1.3B, where the video is generated through 21 autoregressive rollout steps with three latent frames per chunk. The analysis focuses on the self-attention computation and retrieval operations. For a key/value sequence of length $K$, the self-attention FLOPs scale linearly with $K$, as:

\begin{equation}
C_{\mathrm{attn}}(K)=4BHLKd,
\end{equation}

where $B$, $H$, and $d$ denote the number of transformer blocks, attention
heads, and head dimension, respectively.

For full-history attention, the visible memory grows with the rollout length,
resulting in the highest attention cost. Frame-group retrieval reduces this
cost by retrieving a fixed number of latent frames and sharing the retrieved
memory across queries. In contrast, online-clustered retrieval maintains
historical tokens in semantic clusters and performs query-head-wise retrieval.
In our setting, frame-group retrieval selects 20 latent frames.

Table~\ref{tab:efficiency-theory} summarizes the theoretical comparison. The
reported FLOPs include self-attention and retrieval-related computation, while
the collected K/V volume denotes the logical amount of retrieved memory
elements. The theoretical comparison shows that both retrieval strategies
substantially reduce the self-attention cost compared with full-history
attention. With the current configurations, online-clustered retrieval achieves
a slightly lower attention cost than frame-group retrieval by accessing a more
compact set of retrieved tokens. Although the online clustering process
introduces additional computation overhead, this overhead mainly comes from
cluster refinement and maintenance operations, including centroid refinement,
cluster splitting, and merging. Nevertheless, the retrieval overhead remains
small compared with the dominant self-attention computation. The overall cost
therefore remains comparable to frame retrieval while providing finer-grained
and content-adaptive memory access. These results indicate that the proposed
retrieval mechanism improves memory utilization not only through computational
reduction, but also through better alignment with full-memory attention
patterns.

\begin{figure}[!t]
    \begin{center}
        \includegraphics[width=1.0\linewidth]{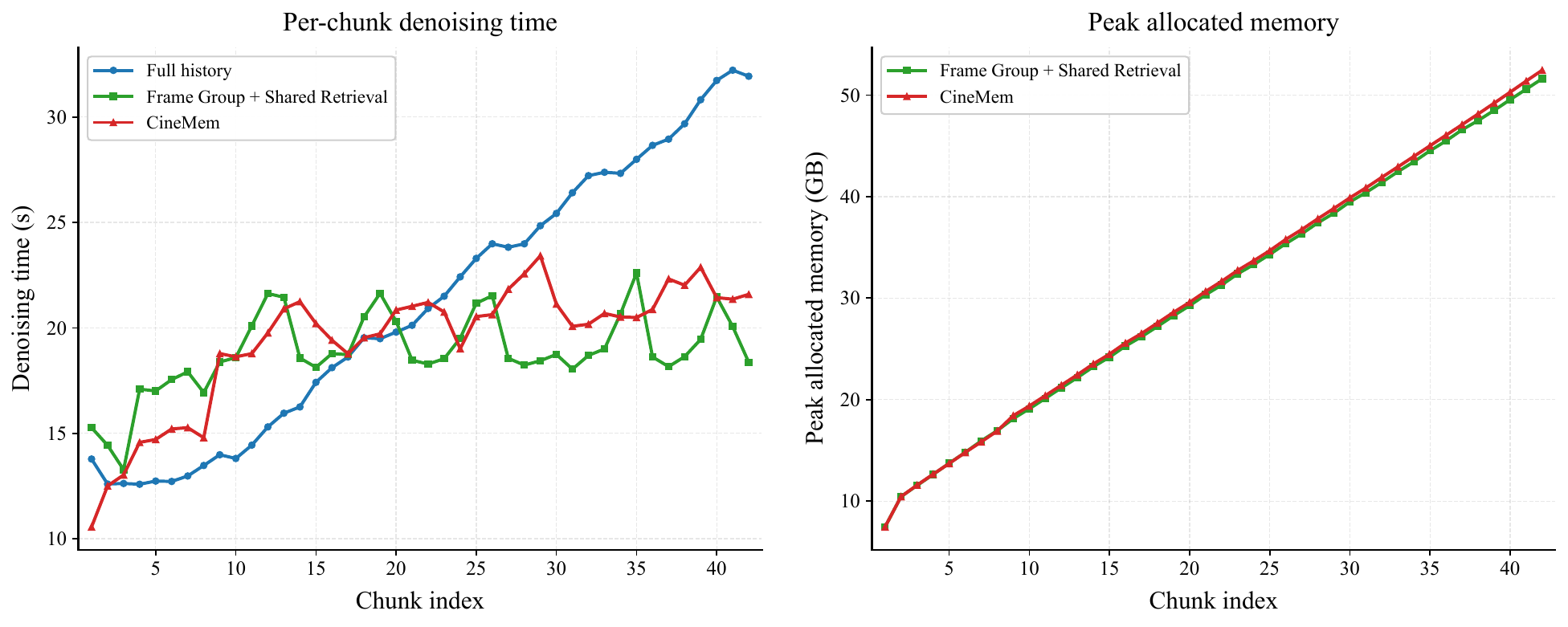  }
    \end{center}
    \caption{\textbf{Practical efficiency comparison during autoregressive rollout.}
The left figure shows the per-chunk denoising time, and the right figure
shows the peak allocated GPU memory during 42-chunk generation.} 
   \label{fig:eff}  
\end{figure}

We further evaluate the practical inference efficiency of different memory
strategies. We measure both the per-chunk denoising time and peak GPU memory
usage during a 42-chunk autoregressive rollout. All measurements are conducted
on four H200 GPUs with sequence parallelism ($sp\_size=4$). As shown in
Figure~\ref{fig:eff}, retrieval-based methods exhibit much
more stable inference time compared with full-history attention. Although
full-history attention achieves lower latency at the beginning of generation,
its denoising time increases continuously as more historical tokens participate
in attention computation. In contrast, both frame-group retrieval and
\methodname{} maintain relatively stable latency by constraining the amount of
 historical memory retrieved. \methodname{} introduces a small additional
overhead compared with frame-group retrieval due to online clustering and
query-adaptive routing, but the difference remains limited throughout the
rollout. 
For peak memory consumption, \methodname{} and frame-group retrieval exhibit
nearly identical memory growth, as shown in Figure~\ref{fig:eff}.
Although \methodname{} maintains additional cluster metadata, the overall memory
usage is dominated by KV cache storage. Consequently, the extra memory
required by cluster statistics and routing information is marginal, resulting
in comparable memory footprints between the two retrieval strategies. These
results demonstrate that \methodname{} achieves fine-grained and adaptive
memory retrieval without introducing substantial practical efficiency
overhead.

%% file: sec/D_evaluation.tex
\section{Details of Evaluation}
\label{sec:sup_evaluation}

In Section \ref{sec:evaluation}, we introduce an evaluation protocol for streaming multi-shot video generation from three perspectives: video-level, intra-shot-level, and inter-shot-level evaluation. In this section, we provide additional details of each evaluation component.

\subsection{Implementation of Metrics}
\label{sec:sup_metric}

For video-level evaluation, we focus on fundamental properties of generated videos, including visual quality, text-video alignment, and transition controllability. We adopt widely used evaluation tools for these aspects, including LAION aesthetic score \citep{LAIONAI}, ViCLIP-based similarity \citep{internvid}, and TransNetV2-based transition evaluation \citep{soucek2024transnetv2}.

For intra-shot-level evaluation, we treat each shot as an individual video and evaluate temporal consistency within the shot. Specifically, we extract frame-level features using DINO \citep{oquab2023dinov2} and compute consistency scores based on feature similarities between frames.

For inter-shot-level evaluation, we first adopt a conventional similarity-based metric commonly used in multi-shot video generation \citep{internvid}. Similar to intra-shot evaluation, this metric measures feature similarity between frames from different shots.
In addition, we introduce a VLM-assisted metric, Prompt-conditioned Transition-Consistency Score (PTCS), which consists of two complementary components: Prompt-conditioned Continuity Preservation (PCP) and Prompt-conditioned Transition Alignment (PTA). Given a hierarchical prompt, we first construct a semantic state graph to identify the semantic elements that should remain unchanged and those expected to change across each transition. PCP evaluates whether the required elements are consistently preserved before and after the transition, while PTA measures whether the intended changes, such as modifications in focal subjects or camera configurations, are correctly realized.


\textbf{PTCS.} For a video containing shots \(\{S_k\}_{k=1}^{K}\), we evaluate each transition boundary \(b\) between \(S_b\) and \(S_{b+1}\). The evaluator takes the middle frames of the two adjacent shots, the global prompt, and their corresponding shot prompts as input. It grounds the prompt-defined semantic elements and returns, for each category \(c\in\{\mathrm{identity},\mathrm{scene},\mathrm{style}\}\), an applicability flag \(a_{b,c}\) and a consistency score \(s_{b,c}\in[0,1]\). Specifically, identity corresponds to subjects and stable appearance, scene corresponds to location and background, and style corresponds to lighting, color treatment, and rendering style.
The continuity score is computed as:
\begin{equation}
    \mathrm{PCP}_b= \frac{1}{|\mathcal C_b|} \sum_{c\in\mathcal C_b}s_{b,c}, \qquad \mathcal C_b=\{c:a_{b,c}=1\},
    \label{eq:ev1}
\end{equation}
Let \(t_b\in[0,1]\) denote the alignment between the prompt-specified transition and the observed compositional change. A required but unobserved transition is assigned \(t_b=0\). If the evaluator identifies \(m_b\) unrequested semantic changes with severities \(e_{b,j}\in[0,1]\), we compute their average severity as:
\begin{equation}
    u_b= \begin{cases} \frac{1}{m_b}\sum_{j=1}^{m_b}e_{b,j}, & m_b>0,\\ 0, & m_b=0. \end{cases}
    \label{eq:ev2}
\end{equation}
The transition alignment score is then defined as:
\begin{equation}
    \mathrm{PTA}_b=t_b(1-u_b).
    \label{eq:ev3}
\end{equation}
Finally, the joint PTCS score is computed as the geometric mean of PCP and PTA:
\begin{equation}
    \mathrm{PTCS}_b=\sqrt{\mathrm{PCP}_b\cdot\mathrm{PTA}_b}.
    \label{eq:ev4}
\end{equation}
This formulation assigns low scores when either unchanged semantic elements are not preserved or intended changes are not correctly realized, as shown in Figure \ref{fig:PTCS}. The reported PCP, PTA, and PTCS values are computed by averaging valid boundary-level scores over all transitions.

\subsection{The effectiveness of Prompt-conditioned Transition-Consistency Score}
\label{sec:sup_PTCS}

As discussed in Section \ref{sec:evaluation}, conventional similarity-based inter-shot consistency metrics evaluate transitions by measuring feature similarity between frames from different shots. However, such metrics inherently favor cases where adjacent shots exhibit limited visual changes, assigning high consistency scores even when the intended transition is not successfully realized. This limitation explains why some methods can achieve high inter-shot consistency scores while exhibiting poor transition control. These observations motivate our design of PTCS, which explicitly considers both semantic preservation and prompt-specified transition changes. Given that PTCS serves as a key metric for evaluating the effectiveness of \methodname{}, we further analyze its validity in this section.

\begin{figure}[!t]
    \begin{center}
        \includegraphics[width=1.0\linewidth]{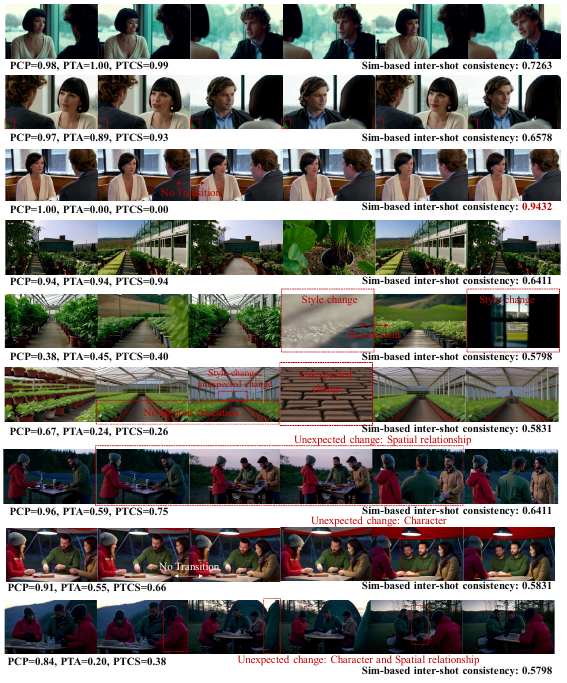  }
    \end{center}
    \caption{\textbf{Qualitative comparison between PTCS and similarity-based inter-shot consistency metrics.} The examples illustrate cases where PTCS better reflects transition correctness by jointly considering semantic preservation and prompt-specified changes.} 
   \label{fig:PTCS_vli}  
\end{figure}

We visualize several multi-shot examples with PTCS evaluations in Figure \ref{fig:PTCS_vli}. Qualitatively, PCP successfully reflects transitions where semantic consistency is violated, while PTA captures both missing transitions, where the intended shot change is not realized, and unexpected changes introduced during transitions, including inconsistencies in style, character identity, spatial relationships, or even subtle local details. By jointly considering these two complementary aspects, PTCS provides scores that better align with the desired properties of shot transitions. In contrast, similarity-based inter-shot consistency metrics may produce misleading evaluations: when the visual composition remains similar, they can assign high scores despite missing or incorrect transitions, while failing to distinguish valid and invalid transitions under substantial compositional changes.

To complement the quantitative evaluation, we conduct a user study to investigate whether PTCS aligns with human perception of inter-shot consistency. Specifically, we select 30 prompts for evaluation, and participants rate each generated video on a scale from 1 to 5 from three perspectives: overall video quality, intra-shot consistency, and inter-shot consistency. The mean opinion scores (MOS) are summarized in Table~\ref{tab:user}.
As shown, the human evaluation results are consistent with the trends observed in PTCS. Methods with higher PTCS scores generally receive higher inter-shot consistency ratings from human participants, while models that achieve high similarity-based consistency scores but fail to perform the intended transitions receive substantially lower human scores. This agreement indicates that PTCS better captures human-perceived transition consistency by jointly considering semantic preservation and prompt-specified changes. Furthermore, \methodname{} achieves the highest MOS in inter-shot consistency and competitive overall video quality, demonstrating its ability to generate coherent multi-shot videos with controllable transitions.

\begin{table}[t]
    \centering
    \scriptsize
    \caption{
    \textbf{User study results.}
    Participants rate generated videos from three perspectives on a scale from 1 to 5.
    The best results are highlighted in \textbf{bold}.
    }
    \label{tab:user}

    \renewcommand{\arraystretch}{1.15}
    \setlength{\tabcolsep}{5pt}

    \begin{tabular}{lccc}
        \toprule
        \textbf{Method}
        &
        \textbf{Overall Video Quality}
        &
        \textbf{Intra-shot Consistency}
        &
        \textbf{Inter-shot Consistency}
        \\
        \midrule

        ShotStream
        & 4.00 $\pm$ 0.63
        & 4.03 $\pm$ 0.77
        & 4.03 $\pm$ 0.42
        \\

        Self-Forcing
        & 2.40 $\pm$ 0.44
        & 3.90 $\pm$ 0.69
        & 1.23 $\pm$ 0.45
        \\

        Infinity-RoPE
        & 2.50 $\pm$ 0.32
        & 3.93 $\pm$ 0.66
        & 1.41 $\pm$ 0.42
        \\

        LongLive2
        & 3.73 $\pm$ 0.50
        & 3.98 $\pm$ 0.80
        & 3.03 $\pm$ 0.50
        \\

        MemFlow
        & 2.71 $\pm$ 0.34
        & 3.93 $\pm$ 0.80
        & 1.97 $\pm$ 0.43
        \\

        HoloCine
        & 4.13 $\pm$ 0.65
        & 4.13 $\pm$ 0.45
        & 4.13 $\pm$ 0.52
        \\

        MultiShotMaster
        & \textbf{4.18 $\pm$ 0.42}
        & 4.07 $\pm$ 0.80
        & 4.30 $\pm$ 0.41
        \\

        \methodname{} (\textbf{Ours})
        & 4.15 $\pm$ 0.47
        & \textbf{4.17 $\pm$ 0.47}
        & \textbf{4.37 $\pm$ 0.31}
        \\

        \bottomrule
    \end{tabular}
\end{table}

%% file: sec/E_Results.tex
\section{Additional Results}
\label{sec:sup_results}

In this section, we provide additional analyses and results to complement the experiments presented in Section \ref{sec:evaluation} and Section \ref{sec:ablation}.

\subsection{Qualitative Results of Ablation Study}

\begin{figure}[!t]
    \begin{center}
        \includegraphics[width=1.0\linewidth]{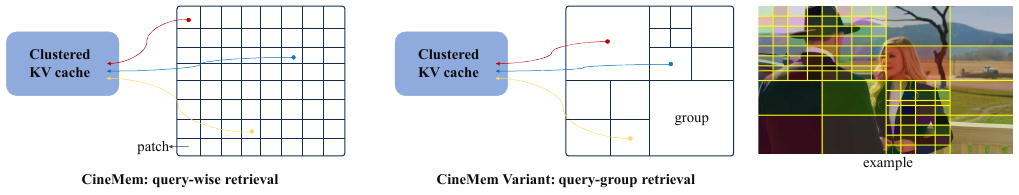  }
    \end{center}
    \caption{Comparison between query-wise retrieval and the query-group retrieval variant. Query-wise retrieval performs independent memory retrieval for each query token, while the query-group variant recursively partitions query tokens based on intra-group content diversity. Queries with similar content preferences are assigned to the same group and share the retrieved memory.} 
   \label{fig:variant}  
\end{figure}

\begin{figure}[!t]
    \begin{center}
        \includegraphics[width=1.0\linewidth]{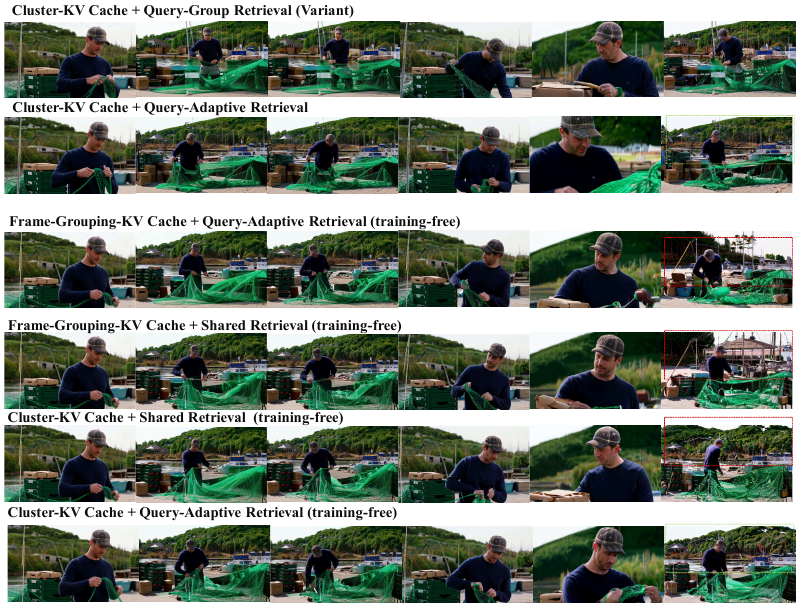  }
    \end{center}
    \caption{A case of qualitative result of ablation study.} 
   \label{fig:ablation}  
\end{figure}

Figure \ref{fig:ablation} provides additional qualitative results for the ablation study. As shown, \methodname{} maintains long-horizon consistency after multiple shot transitions. Even under the training-free setting, our fine-grained retrieval mechanism achieves better results than approaches based on frame-level grouping or shared retrieval. The trained model further improves visual consistency and preserves finer details. For the query-group retrieval variant, we provide additional implementation details in Figure \ref{fig:variant}. This variant also achieves promising performance and serves as an alternative design for efficient query-adaptive retrieval.

\subsection{Fine-Grained Analysis of Generation Quality}
\label{sec:sup_class_res}

In Table~\ref{tab:results}, we present the overall comparison between \methodname{} and existing baselines. In this section, we provide a more fine-grained analysis of the evaluation results by categorizing the evaluation prompts from three perspectives: content category, shot count, and revisit distance.

\begin{figure}[!t]
    \begin{center}
        \includegraphics[width=1.0\linewidth]{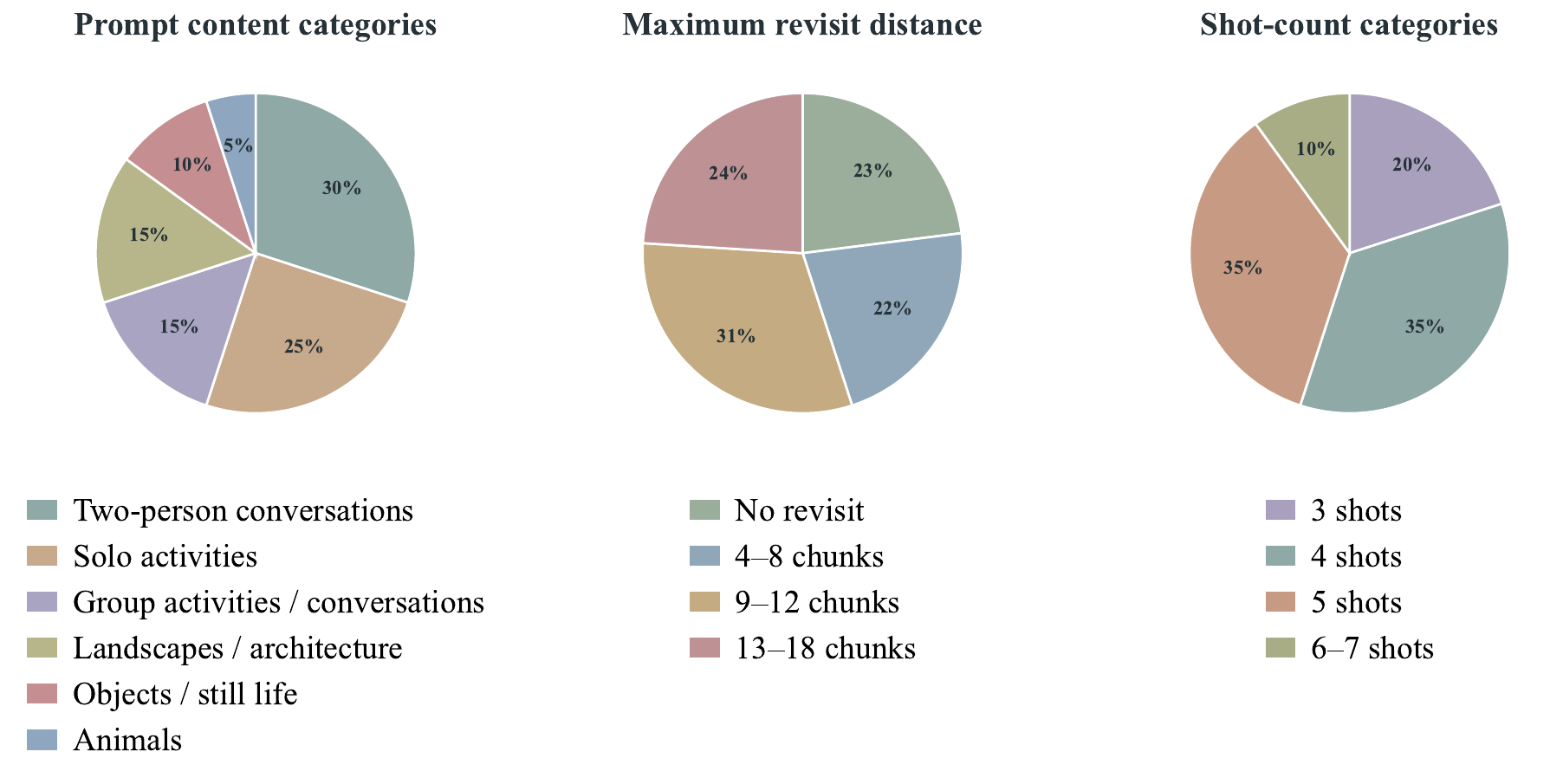  }
    \end{center}
    \caption{Distribution of evaluation prompts across different content categories, shot counts, and revisit distances.} 
   \label{fig:prompt_dis}  
\end{figure}

Figure~\ref{fig:prompt_dis} illustrates the distribution of our evaluation set across these three categories. We use these categorizations to further analyze the performance of different methods under diverse generation scenarios. Here, revisit distance denotes the maximum number of chunks between two revisited appearances within a prompt, reflecting the difficulty of maintaining long-range consistency.

\textbf{Content category.} Table~\ref{tab:content_category} reports the results across six content categories. Overall, \methodname{} exhibits consistently strong transition control and intra-shot consistency across diverse visual content, while achieving the best or second-best PTCS in five out of the six categories. 
In particular, \methodname{} achieves the highest PTCS on \textit{Objects and Still Life} and \textit{Solo Activities}, and remains competitive on more complex scenarios involving conversations, multiple entities, and animals. 
These results suggest that the proposed fine-grained memory retrieval is not specific to a particular type of visual content, but generalizes across substantially different generation scenarios.

\textbf{Shot count.} We further group the evaluation prompts according to the number of shots, as shown in Table~\ref{tab:shot_count}. \methodname{} achieves the highest PTCS across all shot-count groups, while also consistently obtaining the strongest transition control and intra-shot consistency. Notably, its advantage is preserved for prompts containing six to seven shots, where autoregressive error accumulation and long-range dependency modeling become more challenging. This result indicates that our memory retrieval mechanism remains effective as the generation horizon increases.

\textbf{Revisit distance.} Finally, Table~\ref{tab:revisit_distance} analyzes performance with respect to revisit distance, which directly reflects the temporal range over which previously generated content needs to be recalled. 
\methodname{} maintains strong performance across all distance ranges and achieves the highest PTCS for the longest revisit range of 13-18 chunks. 
It also consistently achieves the best transition control and intra-shot consistency across different revisit distances. These results further support the effectiveness of our fine-grained memory retrieval in preserving long-range historical information and maintaining consistency when previously observed content reappears after a long temporal interval.

\begin{table*}[t]
    \centering
    \scriptsize
    \caption{
        \textbf{Fine-grained results across different content categories.}
        The best and runner-up within each category are in \textbf{bold} and \underline{underlined}, respectively.
    }
    \label{tab:content_category}

    \renewcommand{\arraystretch}{1.12}
    \setlength{\tabcolsep}{2.0pt}

    \begin{tabular}{@{}lccccc@{\hspace{4pt}}lccccc@{}}
        \toprule

        \textbf{Method}
        & \textbf{T-Ctrl} $\uparrow$
        & \textbf{T-V Sim.} $\uparrow$
        & \textbf{Aes.} $\uparrow$
        & \textbf{Intra.} $\uparrow$
        & \textbf{PTCS} $\uparrow$
        &
        \textbf{Method}
        & \textbf{T-Ctrl} $\uparrow$
        & \textbf{T-V Sim.} $\uparrow$
        & \textbf{Aes.} $\uparrow$
        & \textbf{Intra.} $\uparrow$
        & \textbf{PTCS} $\uparrow$
        \\

        \midrule

        \multicolumn{6}{c}{\textit{Landscapes and Architecture}}
        &
        \multicolumn{6}{c}{\textit{Objects and Still Life}}
        \\

        \cmidrule(lr){1-6}
        \cmidrule(lr){7-12}

        LongLive2
        & 0.5653 & \textbf{0.2743} & \textbf{0.6086} & 0.9422 & 0.3741
        &
        LongLive2
        & 0.7803 & 0.2444 & 0.5695 & 0.9539 & 0.5017
        \\

        ShotStream
        & \underline{0.9952} & 0.2313 & 0.5639 & 0.9597 & \underline{0.6500}
        &
        ShotStream
        & \textbf{1.0000} & 0.2428 & 0.5579 & 0.9637 & \underline{0.7791}
        \\

        Infinity-RoPE
        & 0.0000 & \underline{0.2674} & 0.5966 & 0.8736 & 0.3743
        &
        Infinity-RoPE
        & 0.1248 & 0.2518 & 0.5692 & 0.9175 & 0.4577
        \\

        Self-Forcing
        & 0.5740 & 0.2606 & 0.5949 & 0.8773 & 0.5071
        &
        Self-Forcing
        & 0.3119 & \underline{0.2551} & \underline{0.5708} & 0.9206 & 0.3354
        \\

        MemFlow
        & 0.2837 & 0.2669 & 0.5846 & 0.8762 & 0.3923
        &
        MemFlow
        & 0.0147 & 0.2424 & \textbf{0.5750} & 0.9453 & 0.3291
        \\

        MultiShotMaster
        & 0.9869 & 0.2204 & 0.5737 & \underline{0.9646} & \textbf{0.7047}
        &
        MultiShotMaster
        & \underline{0.9866} & 0.2545 & 0.5108 & \underline{0.9668} & 0.4981
        \\

        HoloCine
        & 0.9624 & 0.2471 & 0.5377 & 0.9423 & 0.6181
        &
        HoloCine
        & 0.9049 & \textbf{0.2600} & 0.4989 & 0.9377 & 0.6667
        \\

        \methodname{} (\textbf{Ours})
        & \textbf{1.0000} & 0.2434 & \underline{0.6076} & \textbf{0.9777} & 0.6218
        &
        \methodname{} (\textbf{Ours})
        & 0.9680 & 0.2320 & 0.5372 & \textbf{0.9842} & \textbf{0.7841}
        \\

        \midrule

        \multicolumn{6}{c}{\textit{Solo Activities}}
        &
        \multicolumn{6}{c}{\textit{Two-person Conversations}}
        \\

        \cmidrule(lr){1-6}
        \cmidrule(lr){7-12}

        LongLive2
        & 0.4092 & 0.2159 & 0.5598 & 0.9400 & 0.5863
        &
        LongLive2
        & 0.2232 & 0.1875 & 0.5369 & 0.9537 & 0.3356
        \\

        ShotStream
        & 0.9764 & 0.2463 & \underline{0.5718} & \textbf{0.9574} & \underline{0.9241}
        &
        ShotStream
        & 0.7417 & 0.1937 & \textbf{0.6057} & 0.9598 & 0.4784
        \\

        Infinity-RoPE
        & 0.0000 & 0.2233 & 0.5681 & 0.9167 & 0.5320
        &
        Infinity-RoPE
        & 0.0000 & 0.1942 & 0.5745 & 0.9339 & 0.3032
        \\

        Self-Forcing
        & 0.1344 & 0.2372 & 0.5579 & 0.9030 & 0.5639
        &
        Self-Forcing
        & 0.0762 & 0.1948 & 0.5801 & 0.9327 & 0.0862
        \\

        MemFlow
        & 0.0000 & 0.2272 & \textbf{0.5746} & 0.9502 & 0.3303
        &
        MemFlow
        & 0.0000 & 0.1922 & 0.5851 & \underline{0.9604} & 0.0674
        \\

        MultiShotMaster
        & \underline{0.9875} & \underline{0.2579} & 0.5324 & 0.9162 & 0.9031
        &
        MultiShotMaster
        & \underline{0.9669} & \underline{0.2204} & 0.6248 & 0.9484 & \textbf{0.8783}
        \\

        HoloCine
        & \textbf{1.0000} & \textbf{0.2589} & 0.5444 & 0.9211 & 0.8126
        &
        HoloCine
        & 0.9556 & 0.1977 & 0.5718 & 0.9458 & 0.6803
        \\

        \methodname{} (\textbf{Ours})
        & \textbf{1.0000} & 0.2469 & 0.5463 & \underline{0.9514} & \textbf{0.9265}
        &
        \methodname{} (\textbf{Ours})
        & \textbf{0.9977} & \textbf{0.2280} & \underline{0.5992} & \textbf{0.9725} & \underline{0.8682}
        \\

        \midrule

        \multicolumn{6}{c}{\textit{Animals}}
        &
        \multicolumn{6}{c}{\textit{Group Activities and Conversations}}
        \\

        \cmidrule(lr){1-6}
        \cmidrule(lr){7-12}

        LongLive2
        & 0.4415 & 0.2415 & 0.5286 & 0.9494 & 0.4194
        &
        LongLive2
        & 0.2098 & 0.1595 & 0.5425 & 0.9482 & 0.1065
        \\

        ShotStream
        & \textbf{1.0000} & \underline{0.2847} & 0.5312 & \underline{0.9618} & 0.8459
        &
        ShotStream
        & 0.9476 & 0.1853 & 0.5737 & \underline{0.9578} & 0.6633
        \\

        Infinity-RoPE
        & 0.0000 & 0.2628 & 0.5799 & 0.9360 & 0.3415
        &
        Infinity-RoPE
        & 0.0000 & 0.1685 & 0.5664 & 0.9161 & 0.3010
        \\

        Self-Forcing
        & 0.0843 & 0.2575 & 0.5870 & 0.9421 & 0.3074
        &
        Self-Forcing
        & 0.0577 & 0.1706 & 0.5600 & 0.9368 & 0.0404
        \\

        MemFlow
        & 0.0000 & 0.2629 & 0.5640 & 0.9500 & 0.4120
        &
        MemFlow
        & 0.0000 & 0.1630 & \textbf{0.5835} & 0.9492 & 0.0686
        \\

        MultiShotMaster
        & \underline{0.9718} & 0.2817 & 0.5701 & 0.9543 & 0.8536
        &
        MultiShotMaster
        & \underline{0.9832} & 0.1741 & 0.5597 & 0.9196 & 0.7403
        \\

        HoloCine
        & \textbf{1.0000} & \textbf{0.2906} & \textbf{0.5899} & 0.9590 & \textbf{0.9111}
        &
        HoloCine
        & 0.9710 & \underline{0.1995} & 0.5307 & 0.9047 & \textbf{0.8423}
        \\

        \methodname{} (\textbf{Ours})
        & \textbf{1.0000} & 0.2666 & \underline{0.5888} & \textbf{0.9727} & \underline{0.8695}
        &
        \methodname{} (\textbf{Ours})
        & \textbf{1.0000} & \textbf{0.2029} & \underline{0.5817} & \textbf{0.9596} & \underline{0.7796}
        \\

        \bottomrule
    \end{tabular}
\end{table*}

\begin{table*}[t]
    \centering
    \scriptsize
    \caption{
        \textbf{Fine-grained results across different shot counts.}
        The best and runner-up within each category are in \textbf{bold} and \underline{underlined}, respectively.
    }
    \label{tab:shot_count}

    \renewcommand{\arraystretch}{1.12}
    \setlength{\tabcolsep}{2.0pt}

    \begin{tabular}{@{}lccccc@{\hspace{4pt}}lccccc@{}}
        \toprule

        \textbf{Method}
        & \textbf{T-Ctrl} $\uparrow$
        & \textbf{T-V Sim.} $\uparrow$
        & \textbf{Aes.} $\uparrow$
        & \textbf{Intra.} $\uparrow$
        & \textbf{PTCS} $\uparrow$
        &
        \textbf{Method}
        & \textbf{T-Ctrl} $\uparrow$
        & \textbf{T-V Sim.} $\uparrow$
        & \textbf{Aes.} $\uparrow$
        & \textbf{Intra.} $\uparrow$
        & \textbf{PTCS} $\uparrow$
        \\

        \midrule

        \multicolumn{6}{c}{\textit{3 Shots}}
        &
        \multicolumn{6}{c}{\textit{4 Shots}}
        \\

        \cmidrule(lr){1-6}
        \cmidrule(lr){7-12}

        LongLive2
        & 0.3609 & 0.2260 & 0.5578 & 0.9520 & 0.3433
        &
        LongLive2
        & 0.4123 & 0.2000 & 0.5584 & 0.9482 & 0.3866
        \\

        ShotStream
        & 0.9000 & 0.2327 & 0.5669 & \underline{0.9592} & 0.6788
        &
        ShotStream
        & 0.9027 & 0.2170 & \underline{0.5721} & \underline{0.9594} & 0.6750
        \\

        Infinity-RoPE
        & 0.0340 & 0.2379 & \textbf{0.5918} & 0.9251 & 0.4320
        &
        Infinity-RoPE
        & 0.0120 & 0.2081 & 0.5654 & 0.9231 & 0.3781
        \\

        Self-Forcing
        & 0.2502 & 0.2368 & 0.5803 & 0.9346 & 0.2648
        &
        Self-Forcing
        & 0.1966 & 0.2150 & \textbf{0.5751} & 0.9253 & 0.3226
        \\

        MemFlow
        & 0.0680 & 0.2345 & \underline{0.5869} & 0.9517 & 0.1980
        &
        MemFlow
        & 0.0120 & 0.2114 & 0.5699 & 0.9464 & 0.2208
        \\

        MultiShotMaster
        & \underline{0.9770} & \textbf{0.2480} & 0.5709 & 0.9379 & 0.6961
        &
        MultiShotMaster
        & \underline{0.9879} & \underline{0.2292} & 0.5681 & 0.9425 & \underline{0.8582}
        \\

        HoloCine
        & 0.9039 & 0.2431 & 0.5407 & 0.9136 & \underline{0.7164}
        &
        HoloCine
        & 0.9776 & 0.2246 & 0.5319 & 0.9337 & 0.7944
        \\

        \methodname{} (\textbf{Ours})
        & \textbf{0.9840} & \underline{0.2441} & 0.5750 & \textbf{0.9662} & \textbf{0.7516}
        &
        \methodname{} (\textbf{Ours})
        & \textbf{0.9980} & \textbf{0.2303} & 0.5618 & \textbf{0.9634} & \textbf{0.8747}
        \\

        \midrule

        \multicolumn{6}{c}{\textit{5 Shots}}
        &
        \multicolumn{6}{c}{\textit{6--7 Shots}}
        \\

        \cmidrule(lr){1-6}
        \cmidrule(lr){7-12}

        LongLive2
        & 0.3816 & 0.2138 & 0.5570 & 0.9472 & 0.4226
        &
        LongLive2
        & 0.3558 & 0.2179 & 0.5511 & 0.9375 & 0.3861
        \\

        ShotStream
        & 0.9089 & 0.2173 & \textbf{0.5883} & \underline{0.9595} & 0.7164
        &
        ShotStream
        & 0.9397 & 0.2215 & 0.5810 & \underline{0.9592} & 0.6890
        \\

        Infinity-RoPE
        & 0.0000 & 0.2149 & 0.5706 & 0.9095 & 0.3931
        &
        Infinity-RoPE
        & 0.0147 & 0.2214 & \underline{0.5875} & 0.8995 & 0.3178
        \\

        Self-Forcing
        & 0.1339 & 0.2175 & 0.5656 & 0.9089 & 0.2853
        &
        Self-Forcing
        & 0.2095 & 0.2207 & 0.5788 & 0.8797 & 0.3067
        \\

        MemFlow
        & 0.0434 & 0.2112 & 0.5858 & 0.9378 & 0.2179
        &
        MemFlow
        & 0.1102 & 0.2151 & 0.5802 & 0.9170 & 0.3229
        \\

        MultiShotMaster
        & 0.9727 & 0.2216 & 0.5710 & 0.9375 & \underline{0.8192}
        &
        MultiShotMaster
        & \underline{0.9810} & 0.2190 & 0.5726 & 0.9502 & \underline{0.7218}
        \\

        HoloCine
        & \underline{0.9914} & \textbf{0.2332} & 0.5660 & 0.9376 & 0.7018
        &
        HoloCine
        & 0.9725 & \underline{0.2261} & 0.5492 & 0.9510 & 0.7154
        \\

        \methodname{} (\textbf{Ours})
        & \textbf{1.0000} & \underline{0.2328} & \underline{0.5875} & \textbf{0.9693} & \textbf{0.8349}
        &
        \methodname{} (\textbf{Ours})
        & \textbf{1.0000} & \textbf{0.2266} & \textbf{0.6063} & \textbf{0.9757} & \textbf{0.7556}
        \\

        \bottomrule
    \end{tabular}
\end{table*}

\begin{table*}[t]
    \centering
    \scriptsize
    \caption{
        \textbf{Fine-grained results across different revisit distances.}
        The best and runner-up within each category are in \textbf{bold} and \underline{underlined}, respectively.
    }
    \label{tab:revisit_distance}

    \renewcommand{\arraystretch}{1.12}
    \setlength{\tabcolsep}{2.0pt}

    \begin{tabular}{@{}lccccc@{\hspace{4pt}}lccccc@{}}
        \toprule

        \textbf{Method}
        & \textbf{T-Ctrl} $\uparrow$
        & \textbf{T-V Sim.} $\uparrow$
        & \textbf{Aes.} $\uparrow$
        & \textbf{Intra.} $\uparrow$
        & \textbf{PTCS} $\uparrow$
        &
        \textbf{Method}
        & \textbf{T-Ctrl} $\uparrow$
        & \textbf{T-V Sim.} $\uparrow$
        & \textbf{Aes.} $\uparrow$
        & \textbf{Intra.} $\uparrow$
        & \textbf{PTCS} $\uparrow$
        \\

        \midrule

        \multicolumn{6}{c}{\textit{No Revisit}}
        &
        \multicolumn{6}{c}{\textit{4--8 Chunks}}
        \\

        \cmidrule(lr){1-6}
        \cmidrule(lr){7-12}

        LongLive2
        & 0.3696 & 0.2010 & 0.5404 & 0.9472 & 0.3659
        &
        LongLive2
        & 0.3049 & 0.2085 & 0.5411 & 0.9512 & 0.3127
        \\

        ShotStream
        & 0.9609 & 0.2252 & 0.5451 & \underline{0.9603} & 0.7442
        &
        ShotStream
        & 0.7995 & 0.2178 & 0.5856 & \underline{0.9584} & 0.5135
        \\

        Infinity-RoPE
        & 0.0000 & 0.2124 & \underline{0.5542} & 0.9255 & 0.4194
        &
        Infinity-RoPE
        & 0.0000 & 0.2150 & 0.5755 & 0.9232 & 0.3231
        \\

        Self-Forcing
        & 0.1896 & 0.2242 & \textbf{0.5574} & 0.9258 & 0.3223
        &
        Self-Forcing
        & 0.1548 & 0.2176 & 0.5794 & 0.9381 & 0.1502
        \\

        MemFlow
        & 0.0295 & 0.2166 & 0.5511 & 0.9487 & 0.1811
        &
        MemFlow
        & 0.0309 & 0.2138 & \textbf{0.5904} & 0.9547 & 0.1013
        \\

        MultiShotMaster
        & \underline{0.9920} & \underline{0.2364} & 0.5494 & 0.9282 & \textbf{0.8577}
        &
        MultiShotMaster
        & \underline{0.9574} & 0.2275 & \underline{0.5859} & 0.9508 & 0.6240
        \\

        HoloCine
        & 0.9748 & 0.2352 & 0.5233 & 0.9256 & 0.7726
        &
        HoloCine
        & 0.9013 & \underline{0.2336} & 0.5596 & 0.9183 & \underline{0.6249}
        \\

        \methodname{} (\textbf{Ours})
        & \textbf{0.9969} & \textbf{0.2384} & 0.5532 & \textbf{0.9607} & \underline{0.8030}
        &
        \methodname{} (\textbf{Ours})
        & \textbf{0.9854} & \textbf{0.2344} & 0.5848 & \textbf{0.9730} & \textbf{0.8183}
        \\

        \midrule

        \multicolumn{6}{c}{\textit{9--12 Chunks}}
        &
        \multicolumn{6}{c}{\textit{13--18 Chunks}}
        \\

        \cmidrule(lr){1-6}
        \cmidrule(lr){7-12}

        LongLive2
        & 0.4319 & 0.2234 & 0.5642 & 0.9455 & 0.4569
        &
        LongLive2
        & 0.4152 & 0.2101 & 0.5785 & 0.9472 & 0.3995
        \\

        ShotStream
        & 0.9330 & 0.2282 & \underline{0.5844} & \underline{0.9593} & 0.7710
        &
        ShotStream
        & 0.9246 & 0.2094 & \underline{0.5928} & \underline{0.9595} & 0.7024
        \\

        Infinity-RoPE
        & 0.0047 & 0.2285 & 0.5786 & 0.9100 & 0.4633
        &
        Infinity-RoPE
        & 0.0459 & 0.2116 & 0.5888 & 0.9095 & 0.3205
        \\

        Self-Forcing
        & 0.2078 & 0.2225 & 0.5667 & 0.9093 & 0.4032
        &
        Self-Forcing
        & 0.1858 & 0.2184 & 0.5910 & 0.8984 & 0.2676
        \\

        MemFlow
        & 0.0801 & 0.2174 & 0.5726 & 0.9360 & 0.3179
        &
        MemFlow
        & 0.0233 & 0.2170 & \textbf{0.6080} & 0.9296 & 0.2623
        \\

        MultiShotMaster
        & 0.9864 & 0.2312 & 0.5738 & 0.9406 & \textbf{0.8411}
        &
        MultiShotMaster
        & 0.9797 & 0.2217 & 0.5707 & 0.9432 & \underline{0.8467}
        \\

        HoloCine
        & \underline{0.9906} & \underline{0.2324} & 0.5411 & 0.9378 & 0.7754
        &
        HoloCine
        & \underline{0.9900} & \underline{0.2251} & 0.5670 & 0.9465 & 0.7622
        \\

        \methodname{} (\textbf{Ours})
        & \textbf{1.0000} & \textbf{0.2354} & \textbf{0.5858} & \textbf{0.9662} & \underline{0.8228}
        &
        \methodname{} (\textbf{Ours})
        & \textbf{1.0000} & \textbf{0.2259} & 0.5850 & \textbf{0.9697} & \textbf{0.8517}
        \\

        \bottomrule
    \end{tabular}
\end{table*}

\subsection{Longer Video Generation}
\label{sec:longer}

\begin{figure}[t]
  \centering
  \begin{subfigure}{1.0\linewidth}
   \includegraphics[width=1.0\linewidth]{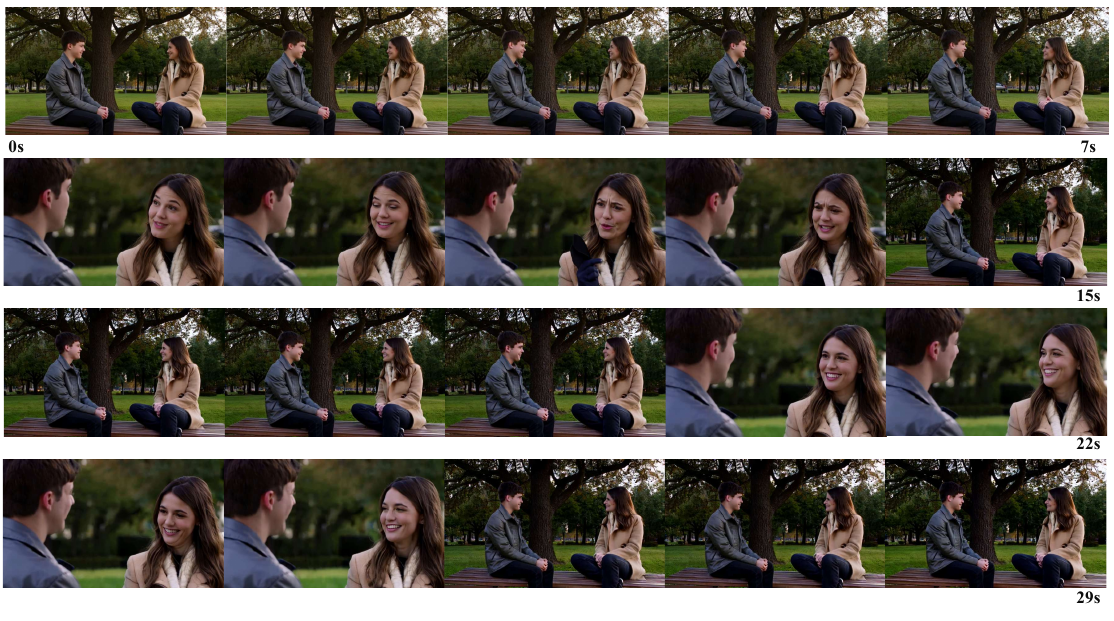}
   \caption{}
   \label{fig:long1}      
  \end{subfigure}
  \hfill
  \begin{subfigure}{1.0\linewidth}
   \includegraphics[width=1.0\linewidth]{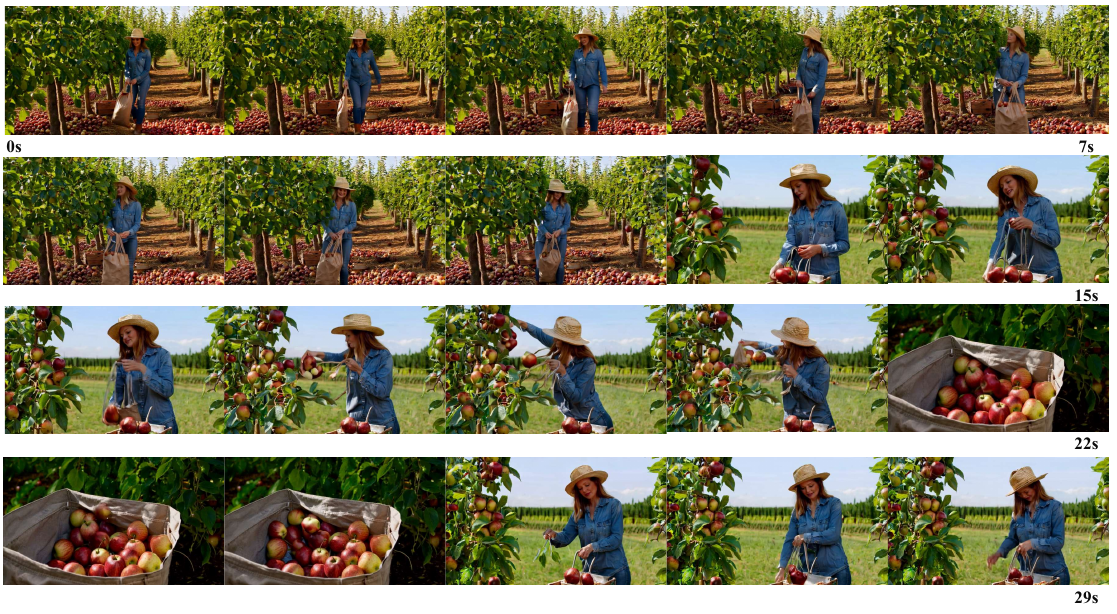}
   \caption{} 
   \label{fig:long2}
  \end{subfigure}
  \caption{Cases of longer video generated by \methodname{}.} 
  \label{fig:long}
\end{figure}

To further investigate the long-horizon generation capability of \methodname{}, we provide additional qualitative results on 30-second videos. As shown in Figure~\ref{fig:long}, \methodname{} can continuously extend autoregressive generation while maintaining coherent visual content and long-range consistency across multiple shot transitions. These results highlight the potential of fine-grained memory retrieval for enabling more scalable autoregressive video generation over extended durations.

%% file: sec/F_limit.tex
\section{Limitation and Failure Case}
\label{sec:limit}

\textbf{Limitations.} Due to limited computational resources, our experiments are conducted on a relatively small-scale autoregressive backbone with a limited amount of training data for adapting a bidirectional model into a long-horizon autoregressive generator. Consequently, the visual quality of generated videos may lag behind that of large-scale state-of-the-art diffusion models with substantially larger model capacity and training data. Nevertheless, our experiments demonstrate the effectiveness of the proposed memory retrieval framework and reveal meaningful insights into memory utilization in autoregressive diffusion models. We expect that the proposed approach can further benefit from larger-scale models and datasets.

\begin{figure}[!t]
    \begin{center}
        \includegraphics[width=1.0\linewidth]{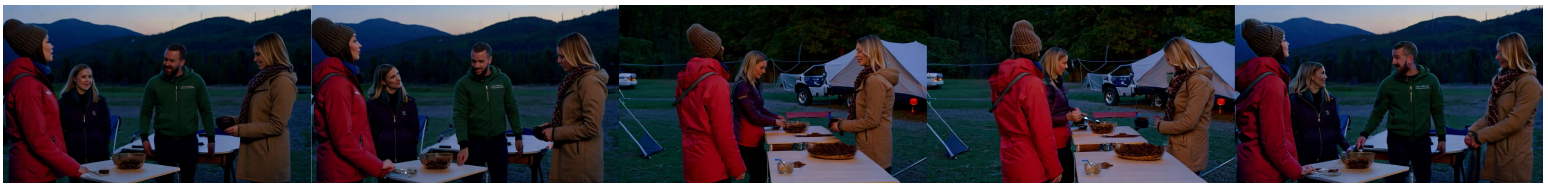}
    \end{center}
    \caption{\textbf{A failure case in complex multi-entity scenarios.} Although the generated video preserves revisitation consistency across shots, inconsistencies such as character disappearance and detail changes occur during the transition from Shot 1 to Shot 2. This example highlights the difficulty of handling multi-entity reasoning and spatial relationship preservation in autoregressive video generation.} 
   \label{fig:case}  
\end{figure}

\textbf{Failure Cases.} Our memory retrieval mechanism aims to approximate the intrinsic full-memory access pattern under a constrained memory budget by enabling more effective historical information retrieval. However, it still struggles in scenarios requiring complex multi-entity reasoning, as illustrated in Figure \ref{fig:case}. In the presented example, the generated video exhibits issues such as missing characters and inconsistent details across shot transitions.
Such failures are particularly challenging for autoregressive generation. In this case, an early chunk introduces four characters through its prompt, while a later chunk requires only three characters. The model must therefore generate a valid transition by adjusting the camera viewpoint while preserving the spatial relationships and identities established in previous shots. However, the earlier chunks are generated without knowledge of future shot requirements, making such coordinated transitions inherently difficult, especially when the initial composition does not provide sufficient flexibility.
Therefore, multi-entity scenarios with complex shot transitions remain challenging even for full-memory autoregressive generation or bidirectional models, as they require reasoning beyond historical information retrieval. While our proposed memory mechanism improves access to relevant historical content and achieves promising results on certain multi-entity scenarios (see supplementary project page), handling complex entity interactions remains an important direction for future work.

%% file: sec/G_future.tex
\section{Future Work}

Beyond the proposed online clustered KV cache and query-adaptive retrieval framework, several directions remain open for future exploration. For memory organization, more effective clustering strategies could further improve retrieval quality. Meanwhile, exploiting the shared retrieval patterns among queries may provide more efficient alternatives to fully query-wise and head-wise retrieval, as explored by the query-group variant in Section \ref{sec:ablation}. These directions may enable more scalable and efficient memory mechanisms for autoregressive video generation.

From the perspective of autoregressive video generation, integrating entity-aware or structure-aware memory retrieval could help address challenging multi-entity scenarios discussed in Section \ref{sec:limit}. Furthermore, extending autoregressive generation to substantially longer horizons remains a promising direction toward more scalable long-form video synthesis.